\documentclass[smallextended]{svjour3}       
\smartqed  
\journalname{Machine Learning}

\usepackage{natbib}
\usepackage{booktabs}
\usepackage{enumerate}
\usepackage{graphicx}
\usepackage{latexsym}
\usepackage{amssymb,amsmath,amsfonts}
\usepackage{color}
\usepackage[ruled,vlined]{algorithm2e}
\usepackage{subcaption}
\usepackage[colorlinks=True,citecolor=blue]{hyperref}
\usepackage{multirow}

\def\urltilda{\kern -.15em\lower .7ex\hbox{\~{}}\kern .04em}

\newtheorem{mydefinition}{Definition}
\newtheorem{myexample}{Example}
\newtheorem{myremark}{Remark}
\newtheorem{mylemma}{Lemma}

\newtheorem{myproposition}{Proposition}
\newenvironment{myproof}{\paragraph{\small{Proof:}}}{\hfill$\Box$}

\newcommand\tab[1][1cm]{\hspace*{#1}}
\newcommand{\Tau}{\mathrm{T}}
\newcommand{\Mu}{\mathrm{M}}

\SetCommentSty{mycommfont}

\begin{document}

\title{Inclusion of Domain-Knowledge into GNNs using Mode-Directed Inverse Entailment}

\titlerunning{Inclusion of Domain-Knowledge into GNNs using MDIE}

\author{Tirtharaj Dash \and
        Ashwin Srinivasan \and \\
        A Baskar
}


\institute{T. Dash \and A. Srinivasan \at
            APPCAIR, BITS Pilani, India
            \and
          T. Dash \and A. Srinivasan \and A. Baskar \at
            Department of CS \& IS \\
            BITS Pilani, K.K. Birla Goa Campus \\
            Goa 403726, India \\
            \email{\{tirtharaj,ashwin,abaskar\}@goa.bits-pilani.ac.in}
}

\date{Received: date / Accepted: date}

\maketitle

\begin{abstract}
We present a general technique for constructing
Graph Neural Networks (GNNs) capable of using multi-relational
domain knowledge. The technique is based on mode-directed inverse entailment
(MDIE) developed in Inductive Logic Programming (ILP). Given a
data instance $e$ and background knowledge $B$, MDIE
identifies a most-specific logical formula
$\bot_B(e)$ that contains all the relational information in $B$
that is related to $e$. We represent
$\bot_B(e)$ by a ``bottom-graph''
that can be converted into a form suitable for GNN implementations.
This transformation allows a
principled way of incorporating generic background knowledge into GNNs: we use the term `BotGNN' for this form of graph neural networks.
For several GNN variants, using real-world datasets
with substantial background knowledge,
we show that
BotGNNs perform significantly better than both
GNNs without background knowledge and a
recently proposed simplified technique for including
domain knowledge into GNNs.
We also provide experimental evidence comparing
BotGNNs favourably to multi-layer perceptrons (MLPs) that use
features representing a ``propositionalised'' form of the background knowledge; and
BotGNNs to a standard ILP based on the use of most-specific
clauses.
Taken together, these results point
	to BotGNNs as capable of combining
	the computational efficacy of GNNs with the representational versatility of
	ILP.
\end{abstract}

\section{Introduction}
\label{sec:intro}

Scientific progress is largely a cumulative enterprise:
hypotheses are devised and experiments conducted
based on what is already known in the field. Recent
ambitious developments (see the Nobel-Turing Grand Challenge~\citep{kitano2016artificial}) seek to accelerate significantly
the process of scientific discovery, by automating
the conjectures-and-refutations part of the scientific approach. At the heart of the approach
is the use of Machine Learning (ML) methods to generate hypotheses about the data
(we are now in the 3rd generation of such ``robot scientists''~\citep{king2009automation},
which are being
used to hypothesise candidate drugs for tropical diseases
like malaria:~\citep{williams2015cheaper}). The use of domain knowledge
is a necessary part of such ML methods. Indeed, a recent
extensive report on AI for Science~\citep{stevens2020ai} has listed
the inclusion of domain knowledge  as the first of 3 Grand
Challenges for ML and AI:

\begin{quote}
``ML and AI are generally domain-agnostic \ldots
Off-the-shelf practice treats [each of these]
datasets in the same way and ignores domain knowledge
that extends far beyond the raw data itself—such as physical
laws, available forward simulations, and established invariances 
and symmetries--that is readily available \ldots
Improving our ability to systematically incorporate
diverse forms of domain knowledge can
impact every aspect of AI, from selection of
decision variables and architecture design to
training data requirements, uncertainty quantification,
and design optimization.''
\end{quote}

ML methods such as Deep Neural Networks (DNNs) have been shown to be extremely
successful at tackling prediction problems across a
range of domains such as image classification~\citep{krizhevsky2017imagenet},
machine translation~\citep{wu2016google}, 
audio generation~\citep{oord2016wavenet}, 
visual reasoning~\citep{johnson2017clevr},
etc.
This has largely been possible due to:
(a) the availability and
easy access to large amount of data, which can be
represented as numeric tensors;
(b) the availability of computational hardware to allow
the massively parallel computations required for large-scale
neural learning.
The capacity to learn from large amounts of vectorised
data---a welcome development in itself---does pose some issues
for a class of real-world problems, which includes
many concerned with scientific discovery. These issues
are:
(a) The available data is structured: often represented
as  graphs of entities and their relationships;
(b) The available data is scarce, with instances
ranging from a few 10s to a few 100s to 1000s. What
is available in compensation, however, is a large amount
of domain-knowledge, that can act as prior information. Examples of such problems abound
in the natural sciences, medicine, and in the
social sciences involving human studies. 

There have been several proposals for a kind of 
DNNs devised specifically to deal with
graph-structured data, called
Graph Neural Networks (GNNs: \citep{wu2020comprehensive}).
The usual approach of \textit{transfer learning} could alleviate the problem of
learning from small amount of data
if the domains of the source- and the target-problem 
are closely related.
The other well-established route for dealing with
small amounts of data involves the use of prior
domain knowledge. Examples of problems with
small amounts of observational data, but with
large amount of compensatory prior-knowledge abound
in the natural sciences, medicine, and in the
social sciences involving human studies.
However, general-purpose ways of incorporating such
knowledge into neural networks remains elusive. In
contrast, Symbolic machine learning techniques such as
Inductive Logic Programming
(ILP: \citep{muggleton1994inductive}) 
have developed generic techniques for incorporating
background knowledge--albeit encoded
as logical statements--into the model-construction process.
This has an immediate importance to problems of scientific
discovery, given the historical focus of scientific disciplines
on mathematical and logical models: as a consequence, a
substantial amount of what is known can be codified in
some logical form. 
Further, the logical representation
used by ILP systems is sufficiently expressive for representing scientific domain knowledge.
However, the subsequent model-construction process can be
computationally expensive, requiring a combinatorial
search that cannot exploit the recent developments in
specialised hardware and software libraries.
In this paper, we adopt the key technique used to
incorporate domain knowledge by one of
the most successful form of ILP, namely, that based on
mode-directed inverse entailment (MDIE: \citep{muggleton1995inverse}).
This form of ILP usually involves a {\em saturation\/}
procedure, which efficiently identifies all the relations
entailed by the domain knowledge for a specific data instance.
This maximal-set of
relations--called a {\em bottom clause\/}--is then used by
an ILP engine to find useful logical explanations for the data.
Here, we develop a corresponding saturation procedure
for GNNs, which results in a maximally-specific
{\em bottom graph\/}, which is then
used for subsequent GNN model construction.
The main contributions of this paper are as follows:
\begin{itemize}
    \item To the field of graph neural networks, the paper proposes a
    systematic technique for
    incorporating symbolic domain-knowledge into GNNs.
    \item To the field of neuro-symbolic learning, 
    the paper provides substantial empirical evidence
    using over 70 real-world datasets and domain-knowledge
    consisting over hundreds of symbolic relations
    that the incorporation of symbolic domain knowledge into
    graph-based neural networks can make a significant
    difference to their predictive performance.
\end{itemize}

The rest of the paper is organised as follows.
We provide a brief specification and a general working
principle of GNNs in Sec.~\ref{sec:gnn}. 
The basic details of saturation as used in MDIE are in Sec.~\ref{sec:mdie}.
Our adaptation of the saturation step to construct
bottom graphs is in Sec.~\ref{sec:botgnn}. Section
\ref{sec:expt} contains an empirical evaluation of
BotGNNs. 
We outline some of the related works in Sec.~\ref{sec:relworks}
providing some relevant methods for incorporating domain-knowledge
into deep neural networks.
Section \ref{sec:concl} concludes the paper.
The Appendices contain some conceptual, implementation- and
application-related details relevant to the content in
the main body of the paper.

\section{Graph Neural Networks (GNNs)}
\label{sec:gnn}

GNNs are a class of deep neural networks suitable
for learning from graph-structured data. GNNs were first
introduced in \citep{gori2005new}, 
and recently being
popularised by their use in molecular property prediction
problems~\citep{gilmer2017neural}.
In this section, we provide a general working principle
of GNNs concerned with graph classification.
We assume that the reader is familiar with the basics of graphs, primarily, directed and undirected graphs,
labelled graphs, the role of a neighbourhood function in a graph, etc.
However, these details are very well discussed in 
\citep[Sec.~2]{dash2021incorporating}.

GNNs are concerned with labelled
graphs, that is, each vertex (and each edge) of
a graph are associated with a numeric feature-vector, 
essentially describing some properties of 
that vertex (or that edge).
Here it suffices to state that the defining
property of a GNN is that it uses some form of
neural message-passing in which messages
(in vector form) are exchanged between vertices
of a graph and updated using a neural network~\citep{gilmer2017neural}. 
For this purpose, a GNN treats the features
associated with a labelled graph as `messages'
and, in the message-passing process, it updates
these messages.
The message-passing process is
conceptually implemented by using a function
called $Relabel$ (this is defined in \citep[Defn.~4]{dash2021incorporating}). This involves
an iterative update of the vertex- (and edge-) labels.
The output of the function is a relabelled graph,
where the vertex- (and edge-) labels are updated.

\begin{figure}[!htb]
    \centering
    \includegraphics[width=0.95\textwidth]{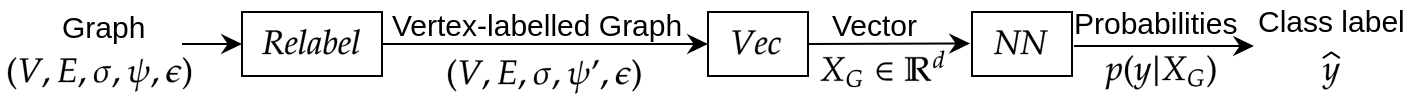}
    \caption{A diagrammatic representation of graph classification using a GNN. Graphs
    are of tuples of the form $(V,E,\sigma,\psi,\epsilon)$, where $V$ is a set of vertices; $E$ is
    a set of edges; $\sigma$ is some neighbourhood function; $\psi$ is a vertex-labelling;
    and $\epsilon$ is an edge-labelling. 
    Often $\sigma$ is left out, and derived from the
    edges in $E$. Also, for GNNs $\psi$ is a mapping from vertices to feature-vectors.
    Many GNN implementations, including the ones used in experiments here, assume
    the graph to be undirected and ignore
    the edge-labelling in $\epsilon$.}
    \label{fig:gnn_block}
\end{figure}

In our present work, we are concerned with problems
involving classification of (molecular) graphs.
This requires a graph-level representation (also
called graph-embedding~\citep{hamilton2020graph}),
meaning, every input graph is encoded as a 
$d$-dimensional feature vector (in $\mathbb{R}^d$). 
This is conceptually implemented using a 
function called $Vec$ (refer \citep[Defn.~5]{dash2021incorporating}) 
that vectorises a
relabelled graph. This feature vector is then
input to a (multi-layered) neural network,
denoted by a function $NN$ that
maps the feature vector to a set of class-labels.
The whole pipeline of graph classification
is shown in Fig.~\ref{fig:gnn_block}.
For completeness of presentation, we now describe how these three functions are related to the general
working principle of GNNs.

\subsection{General working principle of GNNs}
\label{sec:workflow}

Let $G = (V,E,\sigma,\psi,\epsilon)$
denote a graph where $V$ is a set of vertices; $E$ is
a set of edges; $\sigma$ is some neighbourhood function; $\psi$ is a vertex-labelling;
and $\epsilon$ is an edge-labelling.
As mentioned earlier, we are concerned with graph classification
problems. That is, given a graph $G$, a GNN
predicts its class-label. 

In a graph $G$, let $X_v$ denote a vector that represents the initial labelling ($\psi$) of a vertex $v \in V$. That is, 
$X_v$ is the feature-vector associated with the vertex $v$. 
The relabelling function $Relabel:(V,E,\sigma,\psi,\epsilon) \to (V,E,\sigma,\psi',\epsilon)$ 
(iteratively) updates the
labelling of the vertices in $G$. 
This process involves two procedures: 
(a) $\mathsf{AGGREGATE}$: for every
vertex, this procedure aggregates the
information from neighboring vertices;
and (b) $\mathsf{COMBINE}$: this procedure updates the label
of the vertex by combining its present label with
its neighbors'. 
Mathematically, at some iteration
$k$, the labelling of a vertex $v$ (denoted
by $h_v$) is updated as:

\begin{align*}
    a_v^{(k)} &= \mathsf{AGGREGATE}^{(k)}\left(\left\{h_u^{(k-1)}: u \in \mathcal{N}(v)\right\}\right),\\
    h_v^{(k)} &= \mathsf{COMBINE}^{(k)}\left(h_v^{(k-1)},a_v^{(k)}\right)
\end{align*}
where, $\mathcal{N}(v)$ denotes the set of vertices adjacent to $v$. Initially (at $k = 0$), $h_v^{(0)} = X_v$. 

The vectorisation function $Vec:(V,E,\sigma,\psi',\epsilon) \to \mathbb{R}^d$ constructs a vector representation
of the entire graph 
(also called the graph embedding). 
This step is carried out after the representations of all
the vertices are relabelled by some iterations over
$\mathsf{AGGREGATE}$ and $\mathsf{COMBINE}$.
The vectorised representation of the entire graph can be obtained using a $\mathsf{READOUT}$ procedure that aggregates vertex 
features from the final iteration ($k = K$):
\begin{equation*}
    h_G = \mathsf{READOUT}\left(\left\{h_v^{(K)} \mid v \in G \right\}\right)
\end{equation*}

In practice, $\mathsf{AGGREGATE}$ and $\mathsf{COMBINE}$
procedures are implemented using graph convolution
and pooling operations. 
The $\mathsf{READOUT}$ procedure is usually implemented using a global or hierarchical pooling operation~\citep{xu2018how}. 
Variants of GNNs result from modifications to 
these 3 procedures: $\mathsf{AGGREGATE}$,
$\mathsf{COMBINE}$ and $\mathsf{READOUT}$.

\subsection{Note on GNN variants used in this paper}

In our present work, the GNN variants considered
are a result of different graph convolution methods
(we refer the
reader to Appendix~\ref{app:gnnmaths}
for the mathematical aspects of these different approaches):
(1) spectral graph convolution~\citep{Kipf2017gcn},
(2) multistage graph convolution~\citep{morris2019weisfeiler},
(3) graph convolution with attention~\citep{velickovic2018graph},
(4) simple-and-aggregate graph convolution~\citep{hamilton2017inductive},
and (5) graph convolution with auto-regressive
moving average~\citep{bianchi2021graph}.
In addition to the graph convolution methods 
mentioned above, we adopt the method of
graph pooling with
structural-attention~\citep{lee2019self} to apply
down-sampling to graphs. 
We use the hierarchical graph-pooling approach
proposed by \cite{cangea2018towards} to implement
the $\mathsf{READOUT}$ procedure that outputs
a fixed-length representation for an input graph.
This representation is then input to a multilayer
perceptron (MLP) that outputs a class-label.

At this point, the complete architectural
specifics of the GNN including details on various
hyperparameters are not relevant. We
defer these details to Sec.~\ref{sec:meth}.
We refer the reader to Appendix~\ref{app:gnnmaths} 
for a detailed mathematical description on 
the five variants of graph neural networks,
where we describe how
the graph-convolution and graph-pooling methods are
implemented followed by an elaborate 
description on the construction of graph-representation 
using the hierarchical pooling approach.

\section{Mode-Directed Inverse Entailment}
\label{sec:mdie}

Mode-directed Inverse Entailment (MDIE) was introduced
by Stephen Muggleton in~\citep{muggleton1995inverse}, as
a technique for constraining the search for
explanations for data in Inductive Logic Programming
(ILP). For this paper, it is sufficient to
focus on variants of ILP that conforms to
the following input-output requirements:\footnote{
In the following, clauses will usually be in some
subset of first-order logic (usually Horn- or definite-clauses).
When we refer to
a set of clauses, we will usually assume it to be finite.
A set of clauses ${\cal C}$ = $\{D_1,D_2,\dots,D_k\}$ will often be
used interchangebly with the logical formula
${\cal C} = D_1 \wedge D_2 \wedge \dots \wedge D_k$.}

\begin{description}
\item[Given:]
    (i) $B$, a set of clauses (constituting background- or domain-)
    knowledge;
    (ii) a set of clauses $E^+$
	$= \{p_1, p_2, \ldots, p_N\}$ ($N > 0$), denoting
	    a conjunction of ``positive examples''; and
	(iii) a set of clauses $E^{-}
	    = \{\overline{n_1}, \overline{n_2}, \dots, \overline{n_M}\}$ ($M \geq 0$),
	    denoting a conjunction of ``negative examples'', s.t.
	  \begin{description}
	      \item[Prior Necessity.]
	        $B \not \models E^+$
	  \end{description}
\item[Find:] A finite set of clauses (usually
    in a subset of first-order logic), $H$
	$=$  $\{D_1, D_2, \dots, D_k\}$
	s.t.
	\begin{description}
        \item[Weak Posterior Sufficiency.]  For every $D_j \in H$,
            $B \cup \{D_j\} \models p_1\vee p_2\vee \dots \vee p_N$
        \item[Strong Posterior Sufficiency.] $B \cup H \models E^{+}$ 
        \item[Posterior Satisfiability.] $B \cup H \cup E^{-} \not\models \Box$
    \end{description}
\end{description}

\noindent
Here $\models$ denotes logical consequence and $\Box$ denotes
a contradiction.
MDIE implementations attempt to find
the most-probable $H$, given $B$ and the data
$E^+, E^-$.\footnote{
Usually, the entailment relation $\models$ is used to identify
logical consequences of some set
of logical sentences $P$. That is,
what are the $e$'s s.t. $P \models e$?
Here, we are given the $e$'s and $B$, and 
are asking what is an $H$ s.t. $P = B \cup H$. 
In this sense, $H$ is said to be the result of
inverting entailment (IE).}

The key concept used
in \citep{muggleton1995inverse} is to constrain the identification of the
$D_j$ using a {\em most-specific clause\/}. The following is adapted from \citep{muggleton1995inverse}.

\begin{myremark}[Most-Specific Clause]
Given background knowledge $B$ and a data-instance
$e$ (it does not matter at this point
if $e \in E^+$ or $e \in E^-$),
any clause $D$ s.t. $B \cup \{D\} \models e$ will
satisfy $D \models \overline{{B \cup \overline{e}}}$.
This follows directly from the Deduction
Theorem. Let $A = a_1 \wedge  a_2 \cdots \wedge a_n$ be the conjunction of
ground literals\footnote{In theory, the
number of ground literals can be infinite. Practical constraints that restrict
this number to a finite size are described shortly.}
true in all models of ${B \cup \overline{e}}$. 
Hence
${B \cup \overline{e}} \models  a_1 \wedge a_2 \wedge \dots \wedge a_n$.
That is, $\overline{a_1 \wedge a_2 \wedge \dots \wedge a_n}$ $\models$ 
$\overline{{B \cup \overline{e}}}$.
Let $\bot_{B}(e)$ denote
$\overline{a_1 \wedge a_2 \wedge \dots \wedge a_n}$.
That is, $\bot_{B}(e)$ is the clause
$\neg a_1 \vee \neg a_2 \vee \dots \vee \neg a_n $.
Further, since the $a_i$s are ground, $\bot_{B}(e)$ is
a ground clause.

For any clause $D$, if $D \models \bot_{B}(e)$
then $D \models \overline{B \cup \overline{e}}$. Thus
any such $D$ satisfies the
Weak Posterior Sufficiency condition stated earlier.
$\bot_{B}(e)$ is called the most-specific
clause (or ``bottom clause'') for $e$, given $B$.
\end{myremark}

\noindent
Thus, bottom clause construction for a data-instance $e$ provides
a mechanism for inclusion of all the ground
logical consequences given the domain-knowledge $B$ and the
instance $e$.

\begin{myexample}
\label{ex:gparent}
In the following, capitalised letters like
$X,Y$ denote variables. Let
\\\\
\noindent
\begin{minipage}{0.53\linewidth}
${B}$: \\
\tab $parent(X,Y) \leftarrow father(X,Y)$ \\
\tab $parent(X,Y) \leftarrow mother(X,Y)$ \\[6pt]
\tab $mother(jane,alice) \leftarrow$
\end{minipage}%
\begin{minipage}{0.47\linewidth}
${e}$: \\
\tab $gparent(henry,john) ~\leftarrow~$ \\
\tab[2cm] $father(henry,jane)$,\\
\tab[2cm] $mother(jane,john)$ \\
\end{minipage}
\vspace{0.2cm}

\noindent
Here ``$\leftarrow$'' should be read as ``if'' and
the commas (``,'') as ``and''.  So, the
definition for $gparent$ is to be read as:
``henry is a grandparent of john if
henry is the father of jane and
jane is the mother of john.''
\\

\noindent
The conjunction $A$ of
ground literals true in all models of $B \cup \overline{e}$ is:

$\neg gparent(henry,john)$ $\wedge$ 
$father(henry,jane)$ $\wedge$
$mother(jane,john)$ $\wedge$ 
\tab $mother(jane,alice)$ $\wedge$
$parent(henry,jane)$ $\wedge$
$parent(jane,alice)$ $\wedge$ \\
\tab $parent(jane,john)$
\bigskip

\noindent
$\bot_B(e) = \overline{A}$: \\
\tab $gparent(henry,john) \leftarrow$ \\
\tab[2cm] $father(henry,jane)$, $mother(jane,john)$, $mother(jane,alice)$,\\
\tab[2cm] $parent(henry,jane)$, $parent(jane,john)$, $parent(jane,alice)$\\

\noindent
The above clause is logically equivalent to the disjunct:
$gparent(henry,john)$ $\vee~ \neg father(henry,jane)$ $\vee~
\dots \vee~ \neg parent(jane,alice)$.
We will also write clause like this as the set:
$\{$
$gparent(henry,john)$, $\neg father(henry,jane)$,
$\dots$, $\neg parent(jane,alice)$ $\}$.

$\bot_{B}(e)$ thus ``extends'' the example $e$ to
include relations in the background knowledge
provided: our interest is
in the inclusion of the $parent/2$ relation.
The literals in the correct definition of
$gparent$ is a ``generalised'' form of subset of the literals in
$\bot_{B}(e)$. Of course, to find the subset and its generalised form
efficiently is a different
matter, and is the primary concern of ILP systems used to implement
MDIE. 
\end{myexample}

\noindent
It is common to call the non-negated literal in the disjunct
($gparent(henry,john)$) as the ``head''
literal, and the negated literals in the disjunct as the
``body'' literals. In this paper, we will restrict
ourselves to $\bot_B(e)$'s that are definite-clauses 
(clauses with exactly
one head literal). This is for practical reasons, and
not a requirement of the MDIE formulation of most-specific
clauses.

Construction of $\bot_{B}(e)$ is called
a {\em saturation\/} step, reflecting
the extension of the example by all potentially
relevant facts that are derivable using $B$ and
the example $e$. The domain-knowledge can
encode significantly more information than simple
binary relations (like $parent$ above):

\begin{myexample}
Suppose data consist of the atom-and-bond
structure of molecules that are known to be toxic.
Each toxic molecule can be represented by a clausal
formula. For example, a toxic molecule $m_1$
could be represented by the logical formula (here
$a_1$, $a_2$ are atoms,
$c$ denotes carbon, $ar$ denotes aromatic, and so on):

\tab $toxic(m_1) \leftarrow$ \\
\tab[2cm] $atom(m_1,a_1,c)$, \\
\tab[2cm] $atom(m_1,a_2,c)$, \\
\tab[2cm] $\vdots$ \\
\tab[2cm] $bond(m_1,a_1,a_2,ar)$,\\
\tab[2cm] $bond(m_1,a_2,a_3,ar)$, \\
\tab[2cm] $\vdots$
\smallskip

The above clause can be read as: molecule $m_1$ is toxic, if
it contains atom $a_1$ of type carbon, atom $a_2$ of type
carbon, there is an aromatic bond between $a_1$ and $a_2$, and so on.
We will see later that this is a definite-clause encoding
of a graph-based representation of
the molecule $m_1$.

Given background knowledge definitions
(for example, of rings and functional groups), 
$\bot_{B}(e)$ would extend the logical definition
of $e$ with relevant parts of the background knowledge:

\tab $toxic(m_1) \leftarrow$ \\
\tab[2cm] $atom(m_1,a_1,c)$, \\
\tab[2cm] $atom(m_1,a_2,c)$, \\
\tab[2cm] $\vdots$ \\
\tab[2cm] $bond(m_1,a_1,a_2,ar)$,\\
\tab[2cm] $bond(m_1,a_2,a_3,ar)$, \\
\tab[2cm] $\vdots$ \\
\tab[2cm] $benzene(m_1,[a_1,a_2,a_3,a_4,a_5,a_6])$, \\
\tab[2cm] $benzene(m_1,[a_3,a_4,a_8,a_9,a_{10},a_{11}])$, \\
\tab[2cm] $\vdots$ \\
\tab[2cm] $fused(m_1,[a_1,a_2,a_3,a_4,a_5,a_6], [a_3,a_4,a_8,a_9,a_{10},a_{11}])$, \\
\tab[2cm] $\vdots$ \\
\tab[2cm] $methyl(m_1,[\ldots])$, \\
\tab[2cm] $\vdots$
\end{myexample}

\noindent
As seen from this example, the size of $\bot_{B}(\cdot)$ 
can be large. More problematically, for complex
domain knowledge, $\bot_{B}(e)$ may not even be finite.
To address this, MDIE introduces the notion of
a {\em depth-bounded} bottom-clause, using {\em mode}
declarations.

\subsection{Modes}

Practical ILP systems like Progol~\citep{muggleton1995inverse}
use a {\em depth-bounded\/} bottom clause constructed
within a mode-language. We first illustrate a simple
example of a mode-language specification.

\begin{myexample}
\label{ex:modes}
A ``mode declaration'' for an $n$-arity predicate $P$
(often written as $P/n$) is one of the following kinds:
(a) $modeh(P(a_1,a_2,\ldots,a_n))$; or
(b) $modeb(P(a_1,a_2,\ldots,a_n))$.
A set of mode-declarations for the predicates in the
$gparent$ example is:
$\Mu$ = $\{$ 
$modeh(gparent(+person,-person))$,
$modeb(father(+person,-person))$,
$modeb(mother(+person,-person))$,
$modeb(parent(+person,-person))$
$\}$.

The $modeh$ specifies details about literals that can appear
in the head of a clause in the mode-language and
the $modeb$'s  specify details about literals that can appear in
the body of a clause. A ``mode declaration''
refers to either a $modeh$ or $modeb$ statement.
Based on the mode-language specified in
\citep{muggleton1995inverse},
each argument $a_i$ in the mode declarations above is one of:
(1) $+person$, denoting that the argument
    in that literal is an `input' variable of
    type $person$.\footnote{
    Informally, ``a variable of type $\gamma$'' will
    mean that ground substitutions for the variable
    are from some set $\gamma$. Here,
$\gamma$ is the set $person = \{henry,jane,alice,john, \ldots\}$:
that is, $person$ is a unary-relation.}
That is, the variable
must have appeared either as a $-person$ variable
in a literal that appears earlier in the body of the clause or as
a $+person$ variable in the head of the clause; 
(2) $-person$, denoting that the variable in the literal
is an `output' variable of type $person$. 
If an output variable appears in the head of a clause,
it must appear as an output variable of some literal in the body.
There are no special constraint on output variables in body-literals.
That is, they can either be a new variable, or any variable
(of the same type) that has appeared earlier in the clause.
Later we will see how mode-declarations allow the appearance of ground terms.
\end{myexample}

\begin{myexample}
\label{ex:modelang}
Continuing Example \ref{ex:modes},
in the following $X,Y,Z$ are variables
of type $person$.
These clauses are all within the mode language
specified in \citep{muggleton1995inverse}:
(a) $gparent(X,Y) \leftarrow parent(X,Y)$;
(b) $gparent(X,Y) \leftarrow parent(X,X)$;
(c) $gparent(X,Y) \leftarrow mother(X,Y)$; and
(d) $gparent(X,Y) \leftarrow parent(X,Z), parent(Z,Y)$.

But the following clauses are all not within the mode language
in \citep{muggleton1995inverse}:
(e) $gparent(X,Y) \leftarrow parent(Y,Z)$ ($Y$ does not appear before);
(f) $gparent(X,Y) \leftarrow parent(X,Y), parent(Z,Y)$ ($Z$ does not
    appear before);
(g) $gparent(henry,Y) \leftarrow parent(henry,Z), parent(Z,Y)$ ($+$
arguments have to be variables, not ground terms); and
(h) $gparent(X,Y) \leftarrow parent(Z,jane), parent(Z,Y)$ ($-$ arguments have
    to be variables, not ground terms).
\end{myexample}

We refer the reader to \citep{muggleton1995inverse} for
more details on the use of modes. Here we confine ourselves to the
details necessary for the material in this paper.
We first reproduce the notion of a
place-number of a term in a
literal following \citep{plotkin1972automatic}.

\begin{mydefinition}[Term Place-Numbering]
Let $\pi = \langle i_1,\ldots, i_k \rangle$
be a sequence of natural numbers.
We say that a term $\tau$ is in
place-number $\pi$ of a literal $\lambda$ iff: 
(1) $\pi \neq \langle\rangle$; and
(2) $\tau$ is the term at place-number $\langle i_2, \ldots, i_k \rangle$ in
the term at the $i_1^{\mathrm {th}}$ argument of $\lambda$.
$\tau$ is at a place-number $\pi$ in term $\tau'$: 
(1) if $\pi = \langle\rangle$ then
$\tau=\tau'$; and
(2) if $\pi = \langle i_1,\ldots,i_k\rangle$
then $\tau'$ is a term of the form $f(t_1,\ldots,t_m)$, 
$i_1 \leq m$ and $\tau$ is in place-number
$\langle i_2,\ldots,i_k\rangle$ in $t_{i_1}$. 
\label{def:placenum}
\end{mydefinition}



\noindent


\begin{myexample}
\label{ex:placenum}
    {\bf (a)}~In the literal $\lambda = gparent(henry,john)$, the term
$henry$ occurs in the first argument of $\lambda$ and
$john$ occurs in the second argument of $\lambda$. The
place-numbering of $henry$ in $\lambda$ is $\langle 1 \rangle$
and of $john$ in $\lambda$ is $\langle 2 \rangle$.

\noindent
{\bf (b)}~As a more complex example, let $\lambda = mem(a,[a,b,c])$
denote the statement that $a$ is a member of the list $[a,b,c]$.
The second argument of $\lambda$ is short-hand for the term
$list(a,list(b,list(c,nil)))$ (usually, the function
$list/2$ is represented as `.'/2 in the logic-programming literature).
Then the term $a$ is a term that occurs in two place-numbers in $\lambda$:
$\langle 1 \rangle$, and $\langle 2,1 \rangle$. The
term $b$ occurs at place-number $\langle 2,2,1 \rangle$ in $\lambda$; the term $c$ occurs at place-number
$\langle 2,2,2,1 \rangle$ in $\lambda$; and the
term $nil$ occurs at place-number
$\langle 2,2,2,2 \rangle$ in $\lambda$.
\end{myexample}


\noindent
We first present the syntactic aspects
constituting a mode-language.
The meaning of these elements is deferred to the next
section.

\begin{mydefinition}[Mode-Declaration]
\label{def:modedecs}

\begin{enumerate}[(a)]
    \item Let $\Gamma$ be a set of type names. A mode-term  is defined recursively
    as one of:
    (i) $+\gamma$, $-\gamma$ or
    $\#\gamma$ for some $\gamma \in \Gamma$; or
    (ii) $\phi({mt}_1',{mt}_2',\ldots,{mt}_j')$, where
        $\phi$ is a function symbol of arity $j$,
        and the ${mt}_k'$s are mode-terms.
        We will call mode-terms of
        type (i) {\em simple\/} mode-terms and
        mode-declarations of type (ii)
        {\em structured} mode-terms;\footnote{
            For all experiments in this paper, modes
            consist only of simple mode-terms.
        }
\item A mode-declaration $\mu$ is of the form
    $modeh(\lambda')$ or $modeb(\lambda')$.
    Here $\lambda'$ is a ground-literal of
    the form $p({mt}_1,{mt}_2,\ldots,{mt}_n)$ where
    $p$ is a predicate name with arity $n$, and
    the ${mt}_i$ are mode-terms. We will
    say $\mu$ is a $modeh$-declaration
    (resp. $modeb$-declaration)
    for the predicate-symbol $p/n$.\footnote{In general
    there can be several $modeh$ or $modeb$-declarations for
    a predicate-symbol $p/n$. If there is exactly 
    one mode-declaration
    for a predicate symbol $p/n$, we will
    say the mode declaration for $p/n$ is determinate.}
    We will also use $ModeLit(\mu)$ to denote $\lambda'$.
\item $\mu$ is said to be a mode-declaration for a literal
    $\lambda$ iff $\lambda$ and $ModeLit(\mu)$
    have the same predicate
    symbol and arity.
\item Let $\tau$ be the term at place-number $\pi$ in $\mu$,
    We define
    \[ModeType(\mu,\pi) = \left\{\begin{matrix}
    (+, \gamma) & ~\mathrm{if}~ \tau= +\gamma \\
    (-, \gamma) & ~\mathrm{if}~ \tau= -\gamma \\
    (\#, \gamma) & ~\mathrm{if}~ \tau= \#\gamma \\
    unknown      & ~\mathrm{otherwise}
    \end{matrix}\right.\]
\item If $\mu$ is a mode-declaration for literal $\lambda$,
    $ModeType(\mu,\pi)$ = $(+,\gamma)$
    for some place-number $\pi$, 
    $\tau$ is the term at place $\pi$ in
    $\lambda$, 
    then we will say $\tau$ is an input-term
    of type $\gamma$ in $\lambda$ given $\mu$
    (or simply $\tau$ is an input-term of type
    $\gamma$). Similarly we define
    output-terms and constant-terms.
    
\end{enumerate}
\end{mydefinition}

\subsection{Depth-Limited Bottom Clauses}

Returning now to the most-specific clause $\bot_{B}(e)$ for
a data-instance $e$, given background knowledge $B$, it
is sufficient for our purposes to understand that the
input-output specifications in a set of mode-declarations 
result in a natural
notion of the depth at which any
term first appears in $\bot_{B}(e)$
(terms that appear in the head of the
clause are at depth 0, terms that appear
in literals whose input terms depend only
on terms in the head are at depth 1, and so on. A formal
definition follows below.)
By fixing an upper-bound
$d$ on this depth, we can restrict ourselves
to a finite-subset of $\bot_{B}(e)$.\footnote{In
fact, additional restrictions are also needed on
the number of times a relation can occur at any depth.
In implementations like \citep{muggleton1995inverse},
this is usually provided as part
of the mode declaration.}
This is called the {\em depth-limited\/}
bottom clause. Given a set of mode-declarations $\Mu$, we denote
this depth-limited clause by $\bot_{B,\Mu,d}(e)$ (or simply $\bot_d(e)$),
where $d$ is a (pre-specified) depth-limit.
We will refer to the corresponding mode-language
as a depth-limited mode-language and denote
it by ${\cal L}_{\Mu,d}$. We first
illustrate this with an example before defining depth formally.

\begin{myexample}
\label{ex:depthbot}
Using the modes $\Mu$ in Example~\ref{ex:modes},
we obtain the following most-specific clauses for the $gparent$
example (Example~\ref{ex:gparent}):
\vspace{0.2cm}

\noindent
\begin{minipage}{0.5\linewidth}
$\bot_{B,\Mu,1}(e)$: \\
\tab $gparent(henry,john) \leftarrow$ \\
\tab[2cm] $father(henry,jane)$, \\
\tab[2cm] $parent(henry,jane)$ \\ \\ \\ \\
\end{minipage}%
\begin{minipage}{0.5\linewidth}
$\bot_{B,\Mu,2}(e)$: \\
\tab $gparent(henry,john) \leftarrow$ \\
\tab[2cm] $father(henry,jane)$, \\
\tab[2cm] $mother(jane,john)$, \\
\tab[2cm] $mother(jane,alice)$, \\
\tab[2cm] $parent(henry,jane)$, \\
\tab[2cm] $parent(jane,john)$, \\
\tab[2cm] $parent(jane,alice)$
\end{minipage}
\end{myexample}

\noindent

\noindent
We now formally define type-definitions and depth for ground-terms.

\begin{mydefinition}[Type Definitions]
\label{def:types}
Let $\Gamma$ be a set of types and
$\Tau$ be a set of ground-terms. For 
$\gamma \in \Gamma$ we define a set of
ground-terms $T_\gamma$ =
$\{\tau_1,\tau_2,\ldots\}$, where $\tau_i \in \Tau$.
We will say a ground-term
$\tau_i$ is of type $\gamma$ if
$\tau_i \in T_\gamma$, and
denote by $T_\Gamma$ the set
$\{T_\gamma: \gamma \in \Gamma\}$.
$T_\Gamma$ will be called a set of type-definitions.
\end{mydefinition}


\begin{mydefinition}[Depth of a term]
\label{def:termdepth}
Let $\Mu$ be a set of modes.
Let $C$ be a ground clause. Let $\lambda_i$
be a literal in $C$ and let
$\tau$ be an input- or output-term of
type $\gamma$ in $\lambda_i$ given some $\mu \in \Mu$.
Let $Y_\tau$ be the set of
all other terms
in body literals
of $C$ that contain $\tau$
as an output-term of
type $\gamma$. Then,
\[
    depth(\tau) = \left\{\begin{matrix}
		0 & \text{if $\tau$ is an input-term of type $\gamma$}\\
		 & \hfill\text{in a head literal of $C$} \\
		\min_{\tau' \in Y_\tau}depth(\tau') + 1 & \text{otherwise}
	\end{matrix}\right.
\]
\end{mydefinition}

\begin{myexample}
In the previous example for $C = \bot_{B,\Mu,2}(e)$,
$depth(henry)$ = $0$,
$depth(jane)$ = $depth(henry) + 1 = 1$,
$depth(john)$ = $depth(jane) + 1 = 2$, and
$depth(alice)$ = $depth(jane) + 1 = 2$.
\end{myexample}

\noindent
A set of mode-declarations $\Mu$ (see Defn. \ref{def:modedecs}),
a set of type-definitions $T_{\Gamma}$, and a depth-limit $d$
together define a set of acceptable ground
clauses ${\cal L}_{T_{\Gamma},\Mu,d}$.
Informally, ${\cal L}_{T_{\Gamma},\Mu,d}$ consists of ground
clauses in which:
(a) all terms are correctly typed;
(b) all input terms in a body literal have
appeared as output terms in previous body
literals or as input terms in any head literal; 
and
(c) all output terms in any head literal appear
as output terms in some body literals.
In this paper, we will mainly be
interested in definite-clauses (that is,
$m=1$ in the definition that follows).

\begin{mydefinition}[$\lambda\mu$-Sequence]
\label{def:lmpair}
Assume a set of type-definitions
$T_\Gamma$, modes $\Mu$, and a depth-limit
$d$. Let $C=\{l_1,\ldots,l_m,\neg l_{m+1},\ldots,\neg l_k\}$ be a clause with $k$
ground literals.
Then 
$\langle (\lambda_1,\mu_1),(\lambda_2,\mu_2),\dots, (\lambda_k,\mu_k) \rangle$
is said to be a $\lambda\mu$-sequence for $C$ iff it
satisfies the following constraints:
\begin{enumerate}[(a)]
\item 
    (i) The $\lambda$'s are all distinct and 
    (ii) For $j=1 \dots k$, $\mu_j$ is a mode-declaration for $\lambda_j$;
    (iii) For $j=1 \dots m$, $\lambda_j = l_j$ and
        $\mu_j = modeh(\cdot)$;
    (iv) For $j=(m+1)\dots k$, $\lambda_j = l_i$ where $\neg l_i \in C$,
        and $\mu_j = modeb(\cdot)$
     
\item If $\tau$ is an input-term of type $\gamma$
    in $\lambda_j$ given $\mu_j$, then:
    \begin{enumerate}[(i)]
        \item  $\tau \in T_\gamma$; and
        \item  if $j > m$:
        \begin{itemize}
        \item There is an input-term $\tau$ of type
            $\gamma$ in one of $\lambda_1,\dots,\lambda_m$
            given $\mu_1,\dots,\mu_m$; or
        \item There is an output-term $\tau$ of type
            $\gamma$ in $\lambda_i$ ($m < i < j$) given $\mu_i$
        \end{itemize}
    \end{enumerate}
\item If $\tau$ is an output-term of type $\gamma$ in    
        $\lambda_j$ given $\mu_j$, then
        \begin{enumerate}[(i)]
            \item  $\tau \in T_\gamma$; and
            \item  if $j \leq m$:
            \begin{itemize}
                \item $\tau$ is an output-term of type
                $\gamma$ for some $\lambda_i$ ($m < i \leq k$)
                given $\mu_i$
            \end{itemize}
    \end{enumerate}
\item If $\tau$ is a constant-term of type $\gamma$ 
    in $\lambda_j$ given $\mu_j$ then
    $\tau \in T_\gamma$
\item There is no term $\tau$ at any
    place $\pi$ in any $\lambda_j$ s.t.
    the $depth(\tau) > d$.
\end{enumerate}
\end{mydefinition}

\begin{mydefinition}[Mode-Language]
\label{def:modelang}
Assume a set of type-definitions
$T_\Gamma$, modes $\Mu$, and a depth-limit
$d$. The mode-language ${\cal L}_{T_\Gamma,\Mu,d}$ for $T_\Gamma, M, d$ is $\{ C:$  either $C=\emptyset$  or  
there exists a $\lambda\mu$-sequence for $C \}$.

\end{mydefinition}

\noindent

\begin{myexample}
\label{ex:lmpair}
Let $\Mu$ be the set of modes
$\{\mu_1,\mu_2,\mu_3,\mu_4,\mu_5,\mu_6\}$ where
$\mu_1 = modeh(p(+int))$, $\mu_2 = modeh(p(+real))$,
$\mu_3 = modeb(q(+int))$, $\mu_4 = modeb($ $q(+real))$,
$\mu_5 = modeb(r(+int))$, $\mu_6 = modeb(r(+real))$.
Let the depth-limit $d=1$.
Let $C$ be a ground definite-clause $p(1) \leftarrow q(1), r(1)$.
That is, $C = \{p(1), \neg q(1), \neg r(1)\}$.
Let: $\lambda_1 = p(1), \lambda_2 = q(1), \lambda_3 = r(1)$.
Then, $C$ is in ${\cal L}_{\Mu,d}$.
The $\lambda\mu$-sequences for $C$ are:
$\langle (\lambda_1, \mu_1), (\lambda_2, \mu_3), (\lambda_3, \mu_5) \rangle$;
$\langle (\lambda_1, \mu_1), (\lambda_3, \mu_5),$ $(\lambda_2, \mu_3) \rangle$;
$\langle (\lambda_1, \mu_2), (\lambda_2, \mu_4), (\lambda_3, \mu_6) \rangle$;
$\langle (\lambda_1, \mu_2), (\lambda_3, \mu_6), (\lambda_2, \mu_4) \rangle$;

We note that Def.~\ref{def:modelang} does not allow the following to be $\lambda\mu$-sequences:
$\langle (\lambda_1, \mu_1), (\lambda_2, \mu_4), (\lambda_3, \mu_5) \rangle$,
$\langle (\lambda_1, \mu_2), (\lambda_2, \mu_4), (\lambda_3, \mu_5) \rangle$, since
a $1$ of type $int$ is treated as
being different to a $1$ of type $real$.
\end{myexample}

\noindent
We note that although the meanings of
$+$, $-$ and $\#$ are the same here as in 
\citep{muggleton1995inverse}, clauses in
${\cal L}_{T_\Gamma,M,d}$ here are restricted to being
ground (in \citep{muggleton1995inverse},
clauses are required to have variables
in $+$ and $-$ places of literals).

\section{BotGNNs}
\label{sec:botgnn}

In this section, we describe a method to translate the
depth-limited most-specific
clauses of the previous section ($\bot_{B,\Mu,d}(\cdot)$'s)
into a form
that can be used by standard variants of GNNs. We illustrate the
procedure first with an example.

\begin{myexample}
\label{ex:depthbot2}
Consider $\bot_{B,\Mu,2}(e)$ in Example \ref{ex:depthbot}. 
The tabulation below shows the
literals in $\bot_{B,\Mu,2}(e)$ and matching modes

\begin{table}[!htb]
\centering
\begin{tabular}{cll}
    \toprule
    S.No. & \multicolumn{1}{c}{Literal ($\lambda$)} & \multicolumn{1}{c}{Mode ($\mu$)} \\
    \midrule
    1 & $gparent(henry,john)$ & $modeh(gparent(+person,-person))$\\
    2 & $father(henry,jane)$ & $modeb(father(+person,-person))$ \\
    3 & $mother(jane,john)$ & $modeb(mother(+person,-person))$ \\
    4 & $mother(jane,alice)$ & $modeb(mother(+person,-person))$ \\
    5 & $parent(henry,jane)$ & $modeb(parent(+person,-person))$ \\
    6 & $parent(jane,john)$ & $modeb(parent(+person,-person))$ \\
    7 & $parent(jane,alice)$ & $modeb(parent(+person,-person))$ \\
    \bottomrule
\end{tabular}
\end{table}

The table below shows the ground-terms ($\tau$'s) in literals
appearing in $\bot_{B,\Mu,2}(e)$ and their types ($\gamma$'s),
obtained from the
corresponding term-place number in the matching mode:

\begin{table}[!htb]
\centering
\begin{tabular}{cll}
    \toprule
    S.No. & \multicolumn{1}{c}{Term ($\tau$)} & \multicolumn{1}{c}{Type ($\gamma$)} \\
    \midrule
    1 & $henry$ & $person$ \\
    2 & $john$ & $person$ \\
    3 & $jane$ & $person$ \\
    4 & $alice$ & $person$ \\
    \bottomrule
\end{tabular}
\end{table}

The information in these tables can be represented as
a directed bipartite graph as shown in Fig. \ref{fig:exClauseToGraph}.
The square-shaped
vertices represent $(\lambda,\mu)$ pairs in the first table,
and the round-shaped vertices
represent $(\tau,\gamma)$ pairs in the second table.
Arcs from a $(\lambda,\mu)$ (square-) vertex to
a $(\tau,\gamma)$ (round-) vertex indicates term $\tau$ is
designated by mode $\mu$ as an output or constant term
($-$ or $\#$) of type $\gamma$
in literal $\lambda$.
Conversely, an arc from an $(\tau,\gamma)$ vertex to
an $(\lambda,\mu)$ vertex indicates that term $\tau$ is designated
by mode $\mu$ as an input term ($+$) of type $\gamma$ in literal $\lambda$.

\begin{figure}[!htb]
\centering
\begin{subfigure}{.5\textwidth}
  \centering
  \includegraphics[width=.85\linewidth]{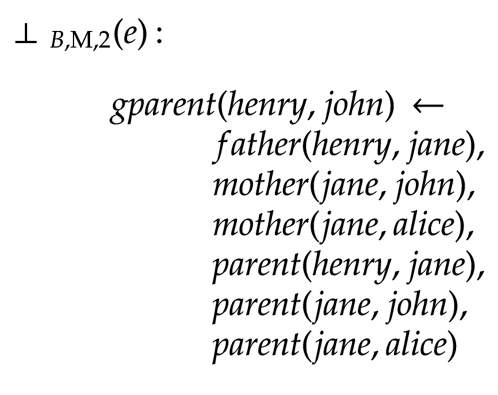}
  \caption{}
  \label{fig:sub1}
\end{subfigure}%
\begin{subfigure}{.5\textwidth}
  \centering
  \includegraphics[width=.7\linewidth]{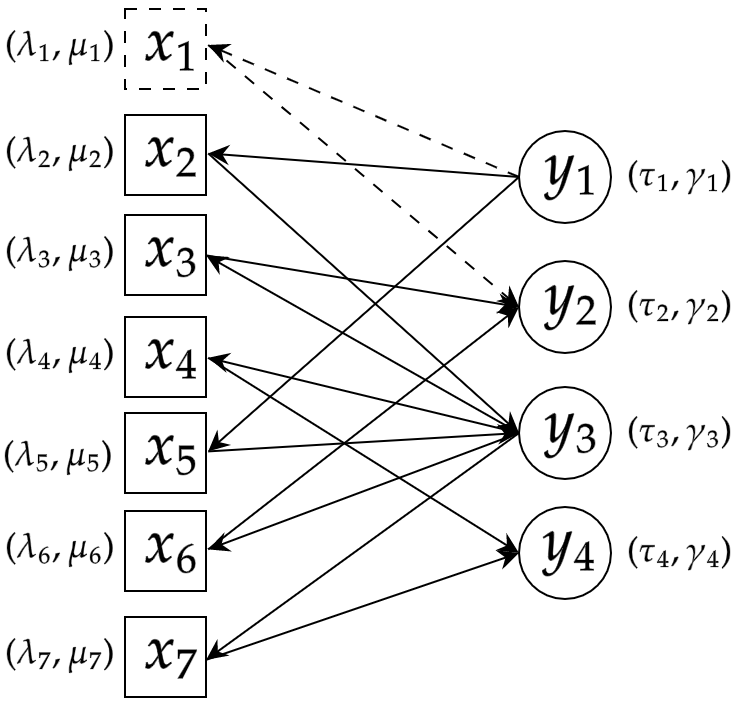}
  \caption{}
  \label{fig:sub2}
\end{subfigure}
\caption{For the $gparent$ example: (a) depth-limited
bottom-clause $\bot_{B,\Mu,2}(e)$; and (b) the
corresponding clause-graph where the vertex-labels
$(\lambda,\mu)$s and $(\tau,\gamma)$s 
are as provided in the preceding tables.
The ``dashed'' square-box and the ``dashed''
arrow are shown to indicate the vertex specifying the
head of the clause.
The subscripts used in the labels correspond to
the S.No. in the tables, for example, $(\lambda_3,\mu_3)$ refers to the third-row in the
first table in this example; and, similarly, 
$(\tau_4,\gamma_4)$ refers to the fourth row in
the second table.
}
\label{fig:exClauseToGraph}
\end{figure}

\end{myexample}
\noindent
The structure in Fig.~\ref{fig:exClauseToGraph} is called a bottom-graph
in this paper.
BotGNNs are GNN models constructed from graphs based
on such clause-graphs.
We first clarify some details needed for the
construction of clause-graphs.

\subsection{Notations and Assumptions}
\label{sec:prelim}

\begin{description}
    \item[Sets.] We use the following notations:
        \begin{enumerate}[(a)]
            \item Set ${\cal E}$ to define a set
            of relational data-instances\footnote{In this paper, this set consists of definite clauses.};
            \item Sets $P, F, K$ to denote predicate-symbols,              function-symbols, and constant-symbols,                respectively;
            \item $\Lambda$ to denote the set of
             all positive ground-literals
            that can be constructed using $P,F,K$; and $\Tau$ to denote
            the set of all ground-terms that can be constructed
            using $F, K$;\footnote{A {\em term\/} is defined recursively as a constant from $K$, a variable, or a function symbol from $F$ applied to term. A ground term is a term without any variables.} 
            \item $\Lambda_C$ to denote the set of all literals in a clause $C$;
            \item $B$ to denote the  set of
                predicate-definitions constituting
                background knowledge;
            \item $\Mu$ to denote the set of modes for
                the predicate-symbols
                in $P$;
            \item Let $\Gamma'$ to denote the set of type-names
                used by modes in $\Mu$. In
                addition, we  assume a
                special type-name $\mathbb{R}$ 
                to denote a numeric type. We denote
                by $\Gamma$ the set $\Gamma'$ $\cup$
               $\{\#t: t \in \Gamma'$ s.t. $\#t$ occurs in some mode $\mu \in \Mu \}$;
        \item $LM$ to denote the set $\{(\lambda,\mu): \lambda \in \Lambda$, $\mu \in \Mu,$ $\mu$ is
        a mode-declaration for $\lambda$ $\}$; and
            $ET$ to denote the set $\{(\tau,\gamma): \tau \in \Tau, \gamma \in \Gamma, \tau ~\mathrm{is~of~type}~\gamma \}$;
        \item $Xs$ to denote the set $\{x_1,\ldots,x_{|LM|}\}$ and
            $Ys$ to denote the set $\{y_1,\ldots,y_{|ET|}\}$;
        \item ${\cal B}$ to denote the set of bipartite
            graphs\footnote{A directed graph $G = (V,E)$ is called bipartite if 
        there is a 2-partition of $V$ into sets $X,Y$, s.t.
        there are no vertices $a,b \in X$ (resp. $Y$) s.t.
        $(a,b) \in E$. We will sometimes denote such a bipartite
        graph by $(X,Y,E)$, where it is understood
        that $V = X ~\cup~ Y$.} of the form 
        $(X,Y,E)$ where $X \subseteq Xs$,
        $Y \subseteq Ys$, and
         $E \subseteq (Xs \times Ys)$ $\cup$ $(Ys \times Xs)$;
        \item  ${\cal G}$ to denote the set of labelled bipartite
        graphs $((X,Y,E),\psi)$ where $(X,Y,E) \in {\cal B}$
        and $\psi: (Xs \cup Ys) \rightarrow (LM \cup ET)$. 
        \item  We will use ${CG}_\top$ to denote the 
        special graph $((\emptyset,\emptyset,\emptyset),\emptyset) \in {\cal G}$.
    \end{enumerate}
    \item[Functions.]  We assume bijections $h_x: LM    \rightarrow Xs$; 
    and $h_y:ET \rightarrow Ys$;
    \item[Implementation.]  We will assume the following
        implementation details:
        \begin{enumerate}[(a)]
            \item  The elements of $\Gamma$ are assumed to be
                unary predicate symbols, and the type-definitions
                $T_\Gamma$ in Defn. \ref{def:types} will be
                implemented as predicate-definitions in
                $B$. That is, if a ground-term $\tau$ is
                of type $\gamma \in \Gamma$ (that
                is $\tau \in T_\gamma$ in Defn. \ref{def:types})
                then $\gamma(\tau) \in B$. We will therefore
                refer to the mode-language
                ${\cal L}_{T_\Gamma,\Mu,d}$ in Defn. \ref{def:modelang}
                as ${\cal L}_{B,\Mu,d}$;
        \item An MDIE implementation that, given
            $B,\Mu,d$, ensures for any ground definite-clause $e$
            returns a unique ground definite-clause
            $\bot_{B,\Mu,d}(e)$ $\in$ ${\cal L}_{B,\Mu,d}$
            if it exists
            or $\emptyset$ otherwise.
            In
            addition, if $\bot_{B,\Mu,d}(e) \in {\cal L}_{B,\Mu,d}$,
            we assume the MDIE implementation has been
            extended to
            return at least one matching $\lambda\mu$-sequence for $\bot_{B,\Mu,d}$.
    \end{enumerate}
\end{description}

\subsection{Construction of Bottom-Graphs}

We now define the graph-structures or simply, the graphs 
constructed from the depth-limited bottom-clauses.
 
\begin{mydefinition}[Literals Set] 
\label{def:litset}
Given background knowledge $B$, a set of modes $\Mu$, a depth-limit $d$,
let $C$ be a clause in ${\cal L}_{B,\Mu,d}$. We define ${Lits}_{B,\Mu,d}(C)$, or
simply $Lits(C)$ as follows:
\begin{enumerate}[(i)]
    \item If $C = \emptyset$ then $Lits(C) = \emptyset$;
    \item If $C \neq \emptyset$,
        let ${\cal{LM}}$       
        be the set of all $\lambda\mu$-sequences for
        $C$.
        Then $Lits(C) = \{(\lambda_i,\mu_i):
    S \in {\cal {LM}} \mathrm{~and~}
    (\lambda_i,\mu_i) \text{~is in sequence~} S \}$
\end{enumerate}
\end{mydefinition}

\noindent
The definition for $Lits(\cdot)$ requires all $\lambda\mu$-sequences
to ensure that $Lits$ is well-defined.
In practice, we restrict ourselves to the 
$\lambda\mu$-sequences identified by the MDIE implementation. 
If these are a subset of all $\lambda\mu$-sequences, 
then the resulting clause-graph will be ``more general'' than that obtained with all $\lambda\mu$-sequences
(see Appendix \ref{app:ClauseToGraphprop}).

\begin{myexample}
We revisit the $gparent$ example.
Let $\Mu$ = $\{\mu_1,\mu_2,\mu_3,\mu_4\}$,
where
$\mu_1 = modeh(gparent(+person,$ $-person))$;
$\mu_2 = modeb(father(+person,-person))$;
$\mu_3 = modeb(mother($ $+person, -person)$;
$\mu_4 = modeb(parent(+person,-person))$.
Let background knowledge $B$ contain
the type-definitions:
$person(henry)$,
$person(john)$,
$person(jane)$, 
$person(alice)$; and
let depth-bound $d=2$,
Let $C = \bot_{B,\Mu,d}(e)$ as in Example~\ref{ex:depthbot}.

\begin{enumerate}
    \item Here $C = \{gparent(henry,john)$,
     $\neg father(henry,jane)$,
     $\neg mother(jane,john)$, 
     $\neg mother(jane,alice)$,
     $\neg parent(henry,jane)$,
     $\neg parent(jane,john)$,
     $\neg parent(jane,alice)\}$.
    \item $\Lambda_C$ = 
        $\{$ $\lambda_1, \lambda_2, \ldots, \lambda_7\}$
        where: $\lambda_1$ = $gparent(henry,john)$,
        $\lambda_2$ = $father(henry,jane)$,
        $\lambda_3$ = $mother(jane,john)$, 
        $\lambda_4$ = $mother(jane,alice)$,
        $\lambda_5$ = $parent(henry,jane)$,
        $\lambda_6$ = $parent(jane,john)$,
        $\lambda_7$ = $parent(jane,alice)$
    \item $C \in {\cal L}_{B,\Mu,d}$ because $S = \langle (\lambda_1, \mu_1)$, $(\lambda_2, \mu_2)$,
    $(\lambda_3, \mu_3)$, $(\lambda_4, \mu_3)$, 
    $(\lambda_5, \mu_4)$, $(\lambda_6, \mu_4)$, 
    $(\lambda_7, \mu_4) \rangle$ is a 
    $\lambda\mu$-sequence for $C$.  Some other permutations of $S$ will also be $\lambda\mu$-sequences.
     The reader can verify that the terms in $\lambda$-components of $S$ 
     are correctly typed; input terms in the body literals appear after corresponding
    output terms in body-literals earlier in the $\lambda$-components of $S$,
    or as input-terms in $\lambda_1$; the output-term in $\lambda_1$
    appears as an output-term in some $\lambda$ later in the sequence $S$.
    \item Then $Lits(C) = \{(\lambda_1,\mu_1)$,
    $(\lambda_2,\mu_2)$, $(\lambda_3, \mu_3)$,
    $(\lambda_4,\mu_3)$, $(\lambda_5,\mu_4)$,
    $(\lambda_6,\mu_4)$, $(\lambda_7,\mu_4) \}$.
 \end{enumerate}
\label{ex:bigexamplelits}
\end{myexample}

\begin{mydefinition}[Terms Set] \label{def:termset}
Given background knowledge $B$, a set of modes $\Mu$, a depth-limit $d$,
let $C \in {\cal L}_{B,\Mu,d}$. We define ${Terms}_{B,\Mu,d}(C)$, or
simply $Terms(C)$ as follows.

If $Lits(C) = \emptyset$, then $Terms(C) = \emptyset$.
Otherwise, for any pair $(\lambda,\mu) \in Lits(C)$,
let
$Ts((\lambda,\mu))$ = $\{(\lambda,\mu,\pi):$
        $\pi$ is a place-number s.t.
        $ModeType(\mu,\pi) = (\cdot,\gamma)$
        for some $\gamma \in \Gamma \}$.
        Then
        $Terms(C) = \bigcup_{x \in Lits(C)}$ $Ts(x)$.
\end{mydefinition}

\begin{myexample}
In Example \ref{ex:bigexamplelits},
$Lits(C)$ =
$\{(\lambda_1,\mu_1)$, $(\lambda_1,\mu_1)$, \dots, $(\lambda_7,\mu_4)\}$.
    Therefore,
    $Terms(C) = \{(\lambda_1,\mu_1, \langle 1 \rangle)$,
    $(\lambda_1,\mu_1, \langle 2 \rangle)$,
    $(\lambda_2,\mu_2, \langle 1 \rangle)$, 
    $(\lambda_2,\mu_2, \langle 2 \rangle)$, 
    \dots
    $(\lambda_7,\mu_4, \langle 1 \rangle)$,
    $(\lambda_7,\mu_4, \langle 2 \rangle)\}$.
\label{ex:bigexampleterms}
\end{myexample}

\begin{mydefinition}[Clause-Graphs]
\label{def:ClauseToGraph}
Given background knowledge $B$, 
a set of modes $\Mu$, and a depth-limit $d$,
we define a function
$ClauseToGraph: {\cal L}_{B,\Mu,d} \rightarrow {\cal G}$ as
follows.

If $C = \emptyset$ then $ClauseToGraph(C) = {CG}_\top$ (see
Sec. \ref{sec:prelim}).
Otherwise, 
$ClauseToGraph(C) = ((X,Y,E),\psi) \in {\cal G}$ where:
\begin{enumerate}[(a)]
\item $X = \{x_i: (\lambda,\mu) \in Lits(C), x_i = h_x((\lambda,\mu))\}$;
\item $Y = \{y_j: (\lambda,\mu,\pi) \in Terms(C)$, $TermType((\lambda,\mu,\pi)) = (\tau,\gamma)$,
                    $ModeType(\mu,\pi) \in \{(+,\gamma),(-,\gamma)\}$,
                    $y_j = h_y((\tau,\gamma))\}$ $\cup$
                    $\{y_j: (\lambda,\mu,\pi) \in Terms(C)$,
$TermType((\lambda,\mu,\pi)) = (\tau,\gamma)$,
                    $ModeType(\mu,\pi) = (\#,\gamma)$,
                    $y_j = h_y((\tau,\#\gamma))\}$;
\item $E ~=~ E_{in} ~\cup~ E_{out}$, where:
    \begin{align*}
        E_{in} &= \{ (y_j,x_i): (\lambda,\mu,\pi)      \in Terms(C),
            x_i = h_x((\lambda,\mu)), \\
            &\tab (\tau,\gamma) = TermType((\lambda,\mu,\pi)),
            y_j = h_y((\tau,\gamma)), \\
            &\tab ModeType(\mu,\pi) = (+,\gamma) \}, \\
        \mathrm{and,~} E_{out} &= \{
            (x_i,y_j): (\lambda, \mu, \pi) \in Terms(C), 
            x_i = h_x((\lambda,\mu)),\\
            &\tab (\tau,\gamma) = TermType((\lambda,\mu,\pi)), 
            y_j = h_y((\tau,\gamma)), \\
            &\tab ModeType(\mu,\pi) \in \{ (-,\gamma),(\#,\gamma) \}
            \}
    \end{align*}
\end{enumerate}
and $\psi$ is a vertex-labelling function defined as follows:
\begin{itemize}
\item[(d)] For $v \in X$, $\psi(v) = h_x^{-1}(v)$;
\item[(e)] For $v \in Y$, $\psi(v) = h_y^{-1}(v)$
\end{itemize}

\end{mydefinition}

\noindent
In Appendix \ref{app:ClauseToGraphprop}, we show
$ClauseToGraph(\cdot)$ is an injective function.

\begin{myexample}
\label{ex:bigexamplecg}
We continue Example \ref{ex:bigexampleterms}.
Recall  $Terms(C)$ = $\{$
        $(\lambda_1,\mu_1,\langle 1 \rangle)$,
        $(\lambda_1,\mu_1,\langle 2 \rangle)$,
        $(\lambda_2,\mu_2,\langle 1 \rangle)$,
        \dots,
        $(\lambda_7,\mu_4,\langle 2 \rangle) \}$.
Then, in Defn. \ref{def:ClauseToGraph},
$TermType((\lambda_1,\mu_1,$ $\langle 1 \rangle))$ =
        $(henry,person)$,
$TermType((\lambda_1,\mu_1,\langle 2 \rangle))$ =
        $(john,person)$,
$TermType($ $(\lambda_2,\mu_2,\langle 1 \rangle))$ =
        $(henry,person)$,
\dots,
$TermType((\lambda_7,\mu_4,\langle 2 \rangle))$ = $(alice,person)$.
Then $ClauseToGraph(C)$ is as follows:
\begin{itemize}
    \item $G = (X,Y,E)$ where:
        \begin{itemize}
            \item $X = \{x_1,x_2,\ldots,x_7\}$,
                where: $x_1 = h_x((\lambda_1,\mu_1))$;
                $x_2 = h_x((\lambda_2,\mu_2))$; \ldots
                $x_7 = h_x((\lambda_7,\mu_4))$
            \item $Y = \{y_1,y_2,y_3,y_4\}$ where:
            $y_1 = h_y((henry,person))$;
            $y_2 = h_y((john,person))$;
            $y_3 = h_y((jane,person))$;
            $y_4 = h_y((alice,person))$
            \item $E = E_{in} \cup E_{out}$, where
            \begin{itemize} 
                \item[]  $E_{in} = 
                    \{
                        (y_1,x_1),
                        (y_1,x_2),
                        (y_1,x_5),
                        (y_3,x_3),
                        (y_3,x_4),
                        (y_3,x_6),
                        (y_3,x_7)
                    \}$
                \item[] $E_{out} =
                    \{
                        (x_1,y_2),
                        (x_2,y_3),
                        (x_3,y_2),
                        (x_4,y_4),
                        (x_5,y_3),
                        (x_6,y_2),
                        (x_7,y_4)
                    \}$
                \end{itemize}
        \end{itemize}
    \item The vertex-labelling $\psi$ is s.t.
            $\psi(x_1)$ = $(\lambda_1,\mu_1)$;
            $\psi(x_2)$ = $(\lambda_2,\mu_2)$;
            $\psi(x_3)$ = $(\lambda_3,\mu_3)$;
            $\psi(x_4)$ = $(\lambda_4,\mu_3)$;
            $\psi(x_5)$ = $(\lambda_5,\mu_4)$;
            $\psi(x_6)$ = $(\lambda_6,\mu_4)$;
            $\psi(x_7)$ = $(\lambda_7,\mu_4)$;
            $\psi(y_1)$ = $(henry,person)$;
            $\psi(y_2)$ = $(john,person)$;
            $\psi(y_3)$ = $(jane,person)$;
            $\psi(y_4)$ = $(alice,person)$.
\end{itemize} 
The reader can compare this to the graph shown
diagrammatically in Fig.~\ref{fig:exClauseToGraph}.
\end{myexample}

    
\begin{myexample}
\label{ex:smallexample}
Examples \ref{ex:bigexamplelits}--\ref{ex:bigexamplecg} 
do not illustrate what happens
when we have multiple matching mode-declarations. To illustrate this
we repeat the exercise with Example~\ref{ex:lmpair} (for consistency, we
now use $\mathbb{R}$ instead of $real$)
In that example,
$\Mu$ =
$\{\mu_1,\mu_2,\mu_3,\mu_4,\mu_5,\mu_6\}$ where
$\mu_1 = modeh(p(+int))$, $\mu_2 = modeh(p(+\mathbb{R}))$,
$\mu_3 = modeb(q(+int))$, $\mu_4 = modeb(q(+\mathbb{R}))$,
$\mu_5 = modeb(r(+int))$, $\mu_6 = modeb(r(+\mathbb{R}))$.
Let the depth-limit $d=1$.

\begin{enumerate}
    \item Here $C$ = $\{$
        $p(1), \neg q(1), \neg r(1)$ $\}$;
    \item $\Lambda_C$ = $\{\lambda_1, \lambda_2, \lambda_3\}$,
    where $\lambda_1 = p(1)$, $\lambda_2 = q(1)$, $\lambda_3 = r(1)$.
    \item $C \in \mathcal{L}_{B,\Mu,d}$ since
    there is at least one $\lambda\mu$-sequence for $C$
    (in fact, there are 4 matching $\lambda\mu$-sequences: see Example~\ref{ex:lmpair}).
    \item $Lits(C) = \{(\lambda_1,\mu_1)$,
    $(\lambda_2, \mu_3)$, $(\lambda_3,\mu_5)$,
    $(\lambda_1, \mu_2)$, $(\lambda_2, \mu_4)$,
    $(\lambda_3, \mu_6)\}$.
    \item We note that the term $1$ is at place-number
    $\langle 1 \rangle$ in all the three literals.
    \item Then $Terms(C) = \{
    (\lambda_1,\mu_1, \langle 1 \rangle)$,
    $(\lambda_2, \mu_3, \langle 1 \rangle)$, \dots,
    $(\lambda_3, \mu_6, \langle 1 \rangle) \}$
    \item Then, in Defn.~\ref{def:ClauseToGraph},
    $TermType((\lambda_1,\mu_1, \langle 1 \rangle))$ = $(1, int)$,
    $TermType((\lambda_2,\mu_3, \langle 1 \rangle))$ = $(1, int)$,
    \ldots,
    $TermType((\lambda_3,\mu_6, \langle 1 \rangle))$ = $(1, \mathbb{R})$.
\end{enumerate}
The reader can verify that $ClauseToGraph(C) = (G,\cdot)$
where $G = (X,Y,E)$ s.t.
\begin{itemize}
    \item $X$ = $\{x_1, x_2, \dots, x_6\}$, where
        $x_1 = h_x((\lambda_1,\mu_1))$, 
        $x_2 = h_x((\lambda_2,\mu_3))$,\dots,
        $x_6 = h_x((\lambda_3,\mu_6))$
    \item $Y = \{y_1, y_2\}$, where 
        $y_1 = h_y((1,int))$ and $y_2 = h_y((1,\mathbb{R}))$
    \item $E$ = $\{(y_1,x_1)$, $(y_1,x_2)$, $(y_1,x_3)$,
        $((y_2,x_4)$, $(y_2,x_5)$, $(y_2,x_6)\}$
\end{itemize}

\end{myexample}

\noindent
It is now straightforward to define graphs from most-specific
clauses.

\begin{mydefinition}[Bottom-Graphs] 
\label{def:botgraph}
Given a data instance $e \in {\cal E}$ and $B,\Mu,d$ as before,
let $\bot_{B,\Mu,d}(e)$
be the (depth-bounded) most-specific ground definite-clause
for $e$. We define
$BotGraph_{B,\Mu,d}(e): {\cal E} \rightarrow {\cal G}$, or
simply $BotGraph(e)$ as follows:
$BotGraph(e) = ClauseToGraph(\bot_{B,\Mu,d}(e))$.
\end{mydefinition}

\begin{myexample}
\label{ex:botgraph_gparent}
For our $gparent/2$ example described through out
this paper, the bottom-graph for the most-specific
clause with $d=2$ is written as:
$BotGraph(e) = ClausetoGraph(\bot_{B,\mu,2}(e))$,
which is shown in the diagram below
(the ``dashed'' square-box and the ``dashed''
arrow are shown to indicate the vertex specifying the
head of the clause):
\begin{center}
    \includegraphics[width=0.35\textwidth]{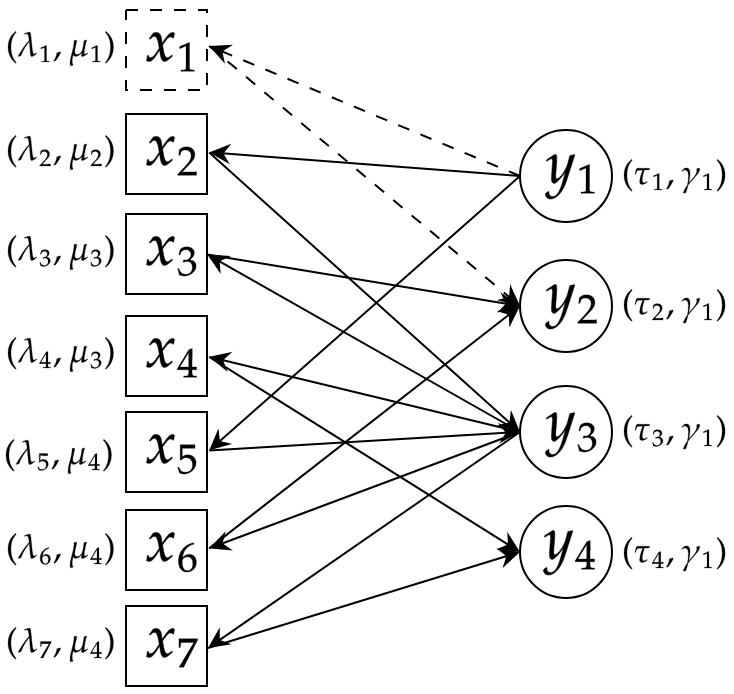}
\end{center}
The vertex-labelling in the above graph is as obtained in Example~\ref{ex:bigexamplecg}, where $\gamma_1$ denotes
the type-name $person$, $\tau_1,\dots,\tau_4$ denote
the terms $henry$, $john$, $jane$, $alice$ respectively.
The reader can verify that the diagram above is consistent
with the bottom-graph shown in Fig.~\ref{fig:exClauseToGraph}.
\end{myexample}

Some properties of clause-graphs are
in Appendix~\ref{app:ClauseToGraphprop}.
The bottom-graphs defined here are not immediately suitable for
GNNs for the task of graph-classification. 
Some graph-transformations are needed before providing them
as input to a GNN. 
We describe these transformations next.



\noindent

\subsection{Transformations for Graph Classification by a GNN}

We now describe functions used to transform bottom-graphs into
a form suitable for the GNN implementations we consider in this paper.
The definite-clause representation of graphs that we use
(an example follows below) contains all the information about the graph
in the antecedent of the definite-clause. The following function
extracts the corresponding parts of the bottom-graph.

\begin{mydefinition}[Antecedent-Graphs]
\label{def:antgraph}
We define function $Antecedent: \mathcal{G} \rightarrow \mathcal{G}$
as follows. Let $(G,\psi) \in \mathcal{G}$, where $G=(X,Y,E)$ is
a directed bipartite graph. 
Let $X_h = \{x: x \in X, \psi(x) = (\lambda, \mu), \mu = modeh(\cdot)\}$.
We define $(G',\psi')$
where $G' = (X',Y',E')$ and
\begin{itemize}
    \item $X' = X - X_h$
    \item $Y' = \{y: y \in Y, \exists x \in X' ~\mathrm{s.t.}~ (x,y)\in E ~\mathrm{or}~ (y,x) \in E\}$
    \item $E' = E - \{(v_i,v_j): v_i \in X_h\}
            - \{(v_j,v_i): v_i \in X_h\}$
\end{itemize}
and, $\psi'(v_i) = \psi(v_i)$
for all $v_i \in X' \cup Y'$.
Then $Antecedent((G,\psi)) = (G',\psi')$.
\end{mydefinition}

Most GNN implementations, including those used in this paper,
require graphs to be undirected~\citep{hamilton2020graph}.
Furthermore, an undirected graph representation allows an 
easy exchange of messages across multiple relations
(the $X$-nodes) resulting in unfolding their
internal dependencies.
We define a function that converts directed clause-graphs to undirected clause-graphs.

\begin{mydefinition}[Undirected Clause-Graphs]
\label{def:ugraph}
We define a function $UGraph: \mathcal{G} \rightarrow \mathcal{G}$
as follows.
Let $(G,\psi) \in \mathcal{G}$, where $G=(X,Y,E)$ is
a directed bipartite graph. We define $(G',\psi')$,
where $G' = (X',Y',E')$ and
\begin{itemize}
    \item $X' = X$
    \item $Y' = Y$
    \item $E' = E \cup \{(v_j,v_i):(v_i,v_j) \in E\}$
\end{itemize}
and $\psi'(v_i) = \psi(v_i)$ for all $v_i \in X' \cup Y'$.
Then $UGraph((G,\psi))$ = $(G',\psi')$. 
\end{mydefinition}

In fact, graphs for GNNs are not actually in ${\cal G}$.
GNN implementations usually require vertices in a graph 
to be labelled with numeric feature-vectors. 
This requires a modification of the vertex-labelling
to be a function from vertices to real-vectors of some finite length.
The final transformation converts the vertex-labelling of
a graph in ${\cal G}$ into a suitable form.

\begin{mydefinition}[Vectorise]
\label{def:vectorise}
Let $(G,\psi) \in \mathcal{G}$, where $G = (X,Y,E)$.
Assume we are given a set of modes $\Mu$.
Let $\Gamma_{\#}$  be the set of all type-names $\gamma \in \Gamma-\{\mathbb{R},\#\mathbb{R}\}$ such
that $\#\gamma$ in some mode $\mu \in \Mu$.
Let $\Tau_{\#} \subseteq \Tau$ be the set of ground-terms
of types in $\Gamma_{\#}$.\footnote{
That is, $\Gamma_{\#}$ is the set of all $\#$-ed, non-numeric type-names in $\Mu$.
and $\Tau_{\#}$ is the set of all ground-terms of $\#$-ed non-numeric
types.}

Let us define the following four functions from $X \cup Y$ to the set of all real vectors of finite length. For $v \in X \cup Y$:
\begin{align*}
f_\rho(v) &= \left\{ 
\begin{matrix}
    onehot(P,r) & \mathrm{~if~} v \in X, 
                  h(v) = (\lambda,\cdot) \mathrm{~and~}
                  predsym(\lambda) = r \\
    \mathbf{0}^{|P|} & \mathrm{~otherwise~}\\
\end{matrix}\right. \\
f_\gamma(v) &= \left\{
\begin{matrix}
    onehot(\Gamma,\gamma) & \mathrm{~if~} v \in Y \mathrm{~and~}
                 h(v) = (\tau,\gamma) \\
    \mathbf{0}^{|\Gamma|} & \mathrm{~otherwise~} \\
\end{matrix}\right. \\
f_\tau(v) &= \left\{
\begin{matrix}
    onehot(\Tau_{\#},\tau) & \mathrm{~if~} v \in Y \mathrm{~and~}
                          h(v) = (\tau,\#\gamma) \mathrm{~and~}
                          \gamma \not \in \mathbb{R} \\
    \mathbf{0}^{|\Tau_{\#}|} & \mathrm{~otherwise~} \\        
\end{matrix}\right. \\
f_\mathbb{R}(v) &= \left\{
\begin{matrix}
    [\tau] & \mathrm{~if~} v \in Y \mathrm{~and~}
                    h(v) = (\tau,\#\mathbb{R}) \\
    \mathbf{0}^{1} & \mathrm{~otherwise~}
\end{matrix} \right.
\end{align*}
where
$\mathbf{0}^{d}$ denotes the zero-vector of length $d$;
$predsym(l)$ is a function that returns the name and arity of
 literal $l$; and $onehot(S,x)$ denotes a one-hot vector encoding of $x \in S$.\footnote{
 A one-hot vector encoding of an element $x$ in a set $S$ assumes a 1-1 mapping $N$ from
 elements of $S$ to $\{1,\ldots,|S|\}$. 
 If $x \in S$ and $onehot(S,x) = {\mathbf v}$ then
 ${\mathbf v}$ is a vector of dimension $|S|$ s.t. $N(x)$'th entry in ${\mathbf v}$ is 1
 and all other entries in ${\mathbf v}$ are 0.
 }
 
Let $Vectorise$ be a function defined on ${\cal G}$
as follows: $Vectorise((G,\psi)) = (G,\psi')$ where $\psi'(v) = f_\rho(v) \oplus f_\gamma(v) \oplus f_\tau(v) \oplus f_\mathbb{R}(v)$ for each $v \in X \cup Y$.
Here  $\oplus$ denotes vector concatenation. 
\end{mydefinition}

\noindent
We note that the vectors in the vertex-labelling from
$Vectorise$ should not be confused with the vector obtained using the $Vec$ function
employed within a GNN (see Fig.~\ref{fig:gnn_block}) in Sec.~\ref{sec:gnn}.
The purpose of that function is to obtain a low-dimensional
real-valued vector representation for an entire graph (usually for
problems of graph-classification).

\begin{myexample}
\label{ex:vectorise1}
Recall the most-specific clause for the $gparent(henry,john)$ in Example~\ref{ex:gparent}:
$gparent(henry,john) \leftarrow$ $father(henry,jane)$, $mother(jane,john)$,
$mother(jane,alice)$, $parent(henry,jane)$,
$parent(jane,john)$, $parent(jane,alice)$. 
The clause-graph and corresponding antecedent-graph
are shown below. 

\noindent
Assume the following sets:
$P = \{gparent/2, father/2, mother/2, parent/2\}$,
$\Gamma = \{person\}$,
$\Gamma_{\#} = \emptyset$,
$\Tau_{\#} = \emptyset$.

Additionally, since the mode-language in Example \ref{ex:gparent} does not have any
$\#$'ed arguments, $\Tau_{\#}= \emptyset$. So:
$f_\rho$ is a 4-dimensional (one-hot encoded)
vector (since $|P| = 4$);
$f_\gamma$ is a 1-dimensional vector (since $|Gamma| = 1$);
$f_\tau$ is a 1-dimensional vector containing $0$
(since $|\Tau_{\#}| = 0$); and
$f_\mathbb{R}$ is a 1-dimensional vector containing $0$
(since there are no $\#$'ed numeric terms)).
A full tabulation of the vectors involved is provided below,
along with the new vertex-labelling that results.
In the table, the vertex labels are as obtained in
Example~\ref{ex:bigexamplecg}; $\gamma_1$ is used
to denote the type $person$.
\begin{center}
\begin{tabular}{ccccccc}
    \hline
    $v$ & $\psi(v)$ & $f_\rho(v)^\top$ & $f_\gamma(v)^\top$ & $f_\tau(v)^\top$ & $f_\mathbb{R}(v)^\top$ & $\psi'(v)^\top$ \\
    \hline
    $x_2$ & $(\lambda_2,\mu_2)$ & $[0,1,0,0]$ & $[0]$ & $[0]$ & $[0.0]$ & $[0,1,0,0,0,0,0.0]$ \\ 
    $x_3$ & $(\lambda_3,\mu_3)$ & $[0,0,1,0]$ & $[0]$ & $[0]$ & $[0.0]$ & $[0,0,1,0,0,0,0.0]$ \\ 
    $x_4$ & $(\lambda_4,\mu_3)$ & $[0,0,1,0]$ & $[0]$ & $[0]$ & $[0.0]$ & $[0,0,1,0,0,0,0.0]$ \\ 
    $x_5$ & $(\lambda_5,\mu_4)$ & $[0,0,0,1]$ & $[0]$ & $[0]$ & $[0.0]$ & $[0,0,0,1,0,0,0.0]$ \\ 
    $x_6$ & $(\lambda_6,\mu_4)$ & $[0,0,0,1]$ & $[0]$ & $[0]$ & $[0.0]$ & $[0,0,0,1,0,0,0.0]$ \\ 
    $x_7$ & $(\lambda_7,\mu_4)$ & $[0,0,0,1]$ & $[0]$ & $[0]$ & $[0.0]$ & $[0,0,0,1,0,0,0.0]$ \\ 
    \hline
    $y_1$ & $(\tau_1,\gamma_1)$ & $[0,0,0,0]$ & $[1]$ & $[0]$ & $[0.0]$ & $[0,0,0,0,1,0,0.0]$ \\
    $y_2$ & $(\tau_2,\gamma_1)$ & $[0,0,0,0]$ & $[1]$ & $[0]$ & $[0.0]$ & $[0,0,0,0,1,0,0.0]$ \\
    $y_3$ & $(\tau_3,\gamma_1)$ & $[0,0,0,0]$ & $[1]$ & $[0]$ & $[0.0]$ & $[0,0,0,0,1,0,0.0]$ \\
    $y_4$ & $(\tau_4,\gamma_1)$ & $[0,0,0,0]$ & $[1]$ & $[0]$ & $[0.0]$ & $[0,0,0,0,1,0,0.0]$ \\
    \hline
\end{tabular}
\end{center}

The following figures show: (a) the antecedent graph and (b) the 
vectorised, undirected, antecedent graph for the $gparent$ example. 
We call the structure in (b) as a BotGNNGraph, the definition of 
which is provided later.
\begin{center}
    \includegraphics[width=0.80\textwidth]{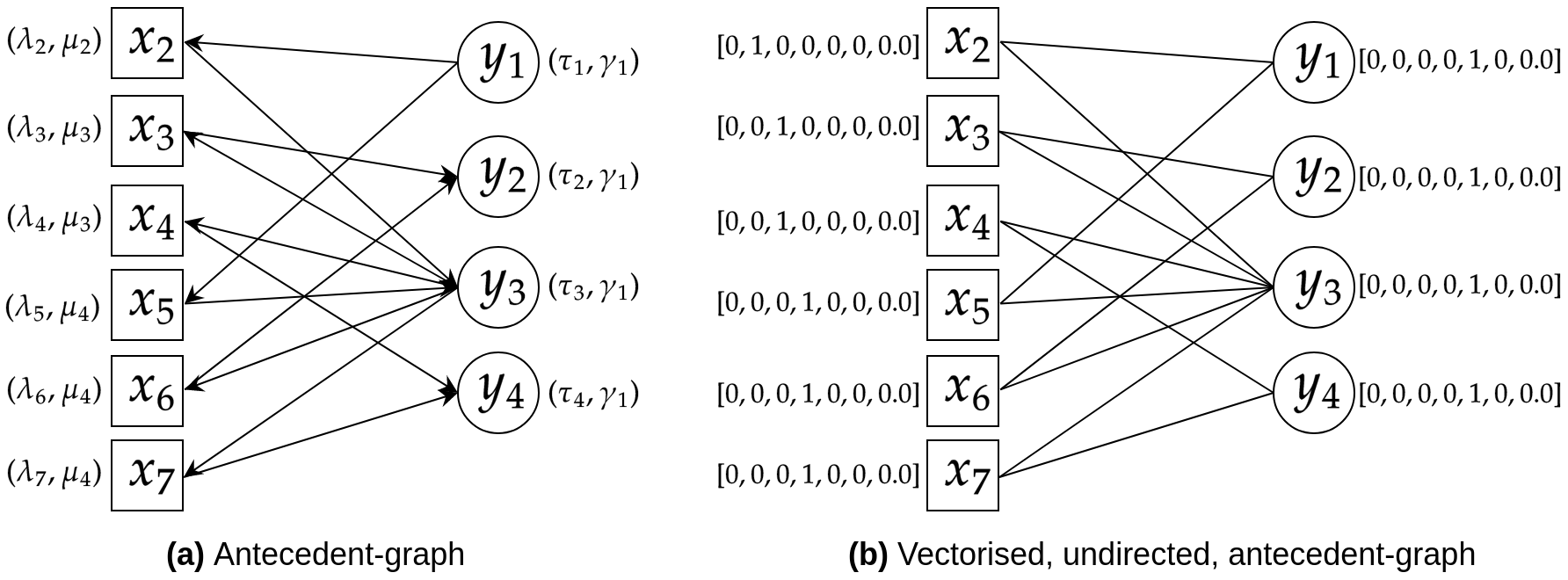}
\end{center}
\end{myexample}

The example above does not have any $\#$-ed arguments
in the modes $\Mu$. In the following example, we 
consider modes that have $\#$-ed arguments (of
types: $\mathbb{R}$ and not $\mathbb{R}$) and
repeat the same exercise: starting with the construction of the bottom-graph.
Then we show how the function $Vectorise$ results
in a vectorised graph suitable for a GNN.

\begin{myexample}
\label{ex:vectorise2}

Let $\Mu$ be the set of modes $\{ \mu_1, \mu_2, \mu_3 \}$
where 
$\mu_1 = modeh(p(+\mathbb{R}))$,
$\mu_2 = modeb(q(+\mathbb{R},\#colour))$,
$\mu_3 = modeb(r(\#colour,\#\mathbb{R}))$.
Let the depth-limit $d = 1$ and
that the background knowledge contains the type-definitions
$colour(white)$ and $colour(black)$.
Let $C$ be a ground definite-clause 
$p(1.0) \leftarrow q(1.0,white), r(white,1.0)$. 
The following are obtained based on the definitions:
\begin{itemize}
    \item $C = \{p(1.0), \neg q(1.0,white), \neg r(white,1.0)\}$.
    \item $\Lambda_C = \{\lambda_1, \lambda_2, \lambda_3\}$,
    where $\lambda_1 = p(1.0)$, $\lambda_2 = q(1.0,white)$,
    $\lambda_3 = r(white,1.0)$. 
    \item $C$ is in $\mathcal{L}_{B,\Mu,d}$ since there is
    at least one $\lambda\mu$-sequence for $C$.
    Here we have one such sequence:
    $\langle (\lambda_1,\mu_1),(\lambda_2,\mu_2),
    (\lambda_3,\mu_3) \rangle$
    \item $Lits(C) = \{(\lambda_1,\mu_1),(\lambda_2,\mu_2),
    (\lambda_3,\mu_3)\}$
    \item $Terms(C) = \{(\lambda_1,\mu_1,\langle 1 \rangle),
    (\lambda_2,\mu_2,\langle 1 \rangle),
    (\lambda_2,\mu_2,\langle 2 \rangle),
    (\lambda_3,\mu_3,\langle 1 \rangle)
    (\lambda_3,\mu_3,\langle 2 \rangle)
    \}$
    \item $TermType((\lambda_1,\mu_1,\langle 1 \rangle)) = (1.0,\mathbb{R})$, 
    $TermType((\lambda_2,\mu_2,\langle 1 \rangle)) = (1.0,\mathbb{R})$,
    $TermType((\lambda_2,\mu_2,\langle 2 \rangle)) = (white,\#colour)$,
    $TermType((\lambda_3,\mu_3,\langle 1 \rangle)) = (1.0,\#\mathbb{R})$
    $TermType((\lambda_3,\mu_3,\langle 2 \rangle)) = (white,\#colour)$
\end{itemize}
Then, $ClauseToGraph(C) = (G,\psi)$, where $G = (X,Y,E)$ s.t.
\begin{itemize}
    \item $X = \{x_1, x_2, x_3 \}$, where 
    $x_1 = h_x((\lambda_1,\mu_1))$, 
    $x_2 = h_x((\lambda_2,\mu_2))$ and
    $x_3 = h_x((\lambda_3,\mu_3))$
    \item $Y = \{y_1, y_2, y_3\}$, where
    $y_1 = h_y((1.0,\mathbb{R}))$, 
    $y_2 = h_y((white,\#colour))$,
    $y_3 = h_y((1.0,\#\mathbb{R}))$
    \item $E = \{(y_1,x_1), (y_1,x_2), (x_2,y_2), (x_3,y_2), (x_3,y_3)\}$
\end{itemize}
and, the vertex-labelling $\psi$ is as follows: 
    $\psi(x_1) = (\lambda_1,\mu_1)$, 
    $\psi(x_2) = (\lambda_2,\mu_2)$, 
    $\psi(x_3) = (\lambda_3,\mu_3)$,
    $\psi(y_1) = (1.0,\mathbb{R})$, 
    $\psi(y_2) = (white,\#class)$,
    $\psi(y_3) = (1.0,\#\mathbb{R})$.
    
\noindent
In this example, we assume the following sets:
$P = \{p/1, q/2, r/2\}$,
$\Gamma = \{\mathbb{R}, \#colour, \#\mathbb{R}\}$,
$\Gamma_{\#} = \{\#colour\}$,
$\Tau_{\#} = \{white,black\}$.

The graph $(G,\psi)$ constructed above is the bottom-graph 
for this particular example. 
The feature-vectors obtained from the functions in $Vectorise$
are tabulated below. In the table, $\tau_1 = 1.0$, $\tau_2 = white$,
$\gamma_1 = \mathbb{R}$, $\gamma_2 = \#colour$, $\gamma_3 = \#\mathbb{R}$.

\begin{center}
\begin{tabular}{ccccccc}
    \hline
    $v$ & $\psi(v)$ & $f_\rho(v)^\top$ & $f_\gamma(v)^\top$ & $f_\tau(v)^\top$ & $f_\mathbb{R}(v)^\top$ & $\psi'(v)^\top$ \\
    \hline
    $x_2$ & $(\lambda_2,\mu_2)$ & $[0,1,0]$ & $[0,0,0]$ & $[0,0]$ & $[0.0]$ & $[0,1,0,0,0,0,0,0,0.0]$ \\
    $x_3$ & $(\lambda_3,\mu_3)$ & $[0,0,1]$ & $[0,0,0]$ & $[0,0]$ & $[0.0]$ & $[0,0,1,0,0,0,0,0,0.0]$ \\
    \hline
    $y_1$ & $(\tau_1,\gamma_1)$ & $[0,0,0]$ & $[1,0,0]$ & $[0,0]$ & $[0.0]$ & $[0,0,0,1,0,0,0,0,0.0]$ \\ 
    $y_2$ & $(\tau_2,\gamma_2)$ & $[0,0,0]$ & $[0,1,0]$ & $[1,0]$ & $[0.0]$ & $[0,0,0,0,1,0,1,0,0.0]$ \\
    $y_3$ & $(\tau_1,\gamma_3)$ & $[0,0,0]$ & $[0,0,1]$ & $[0,0]$ & $[1.0]$ & $[0,0,0,0,0,1,0,0,1.0]$ \\
    \hline
\end{tabular}
\end{center}

\noindent
The following figure shows how the final vectorised graph is constructed
from the bottom-graph (the dotted square-box and the dotted
arrow are shown to indicate the vertex specifying the
head of the clause $C$):
\begin{center}
    \includegraphics[width=0.80\textwidth]{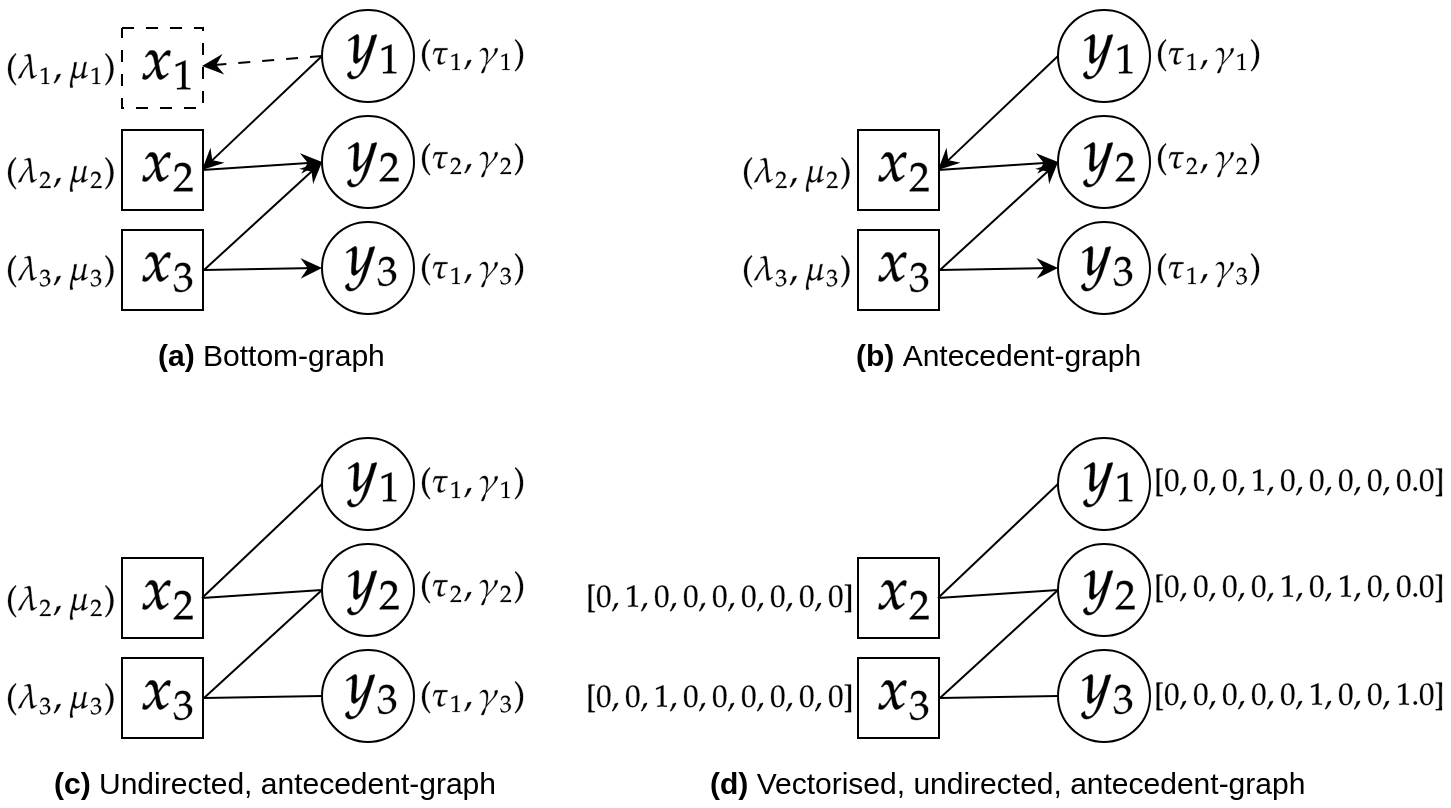}
\end{center}

\end{myexample}

\noindent
The functions $Antecedent$, $UGraph$ and $Vectorise$ transform bottom-graphs into
a form suitable for GNNs by straightforward composition:

\begin{mydefinition}[Graph Transformation]
\label{def:transformgraph}
We define a transformation over ${\cal G}$ as follows:
$TransformGraph(G) = Vectorise(UGraph(Antecedent((G,\psi))))$.
\end{mydefinition}

\noindent
We now have all the pieces for obtaining graphs suitable for GNNs:

\begin{mydefinition}[BotGNN Graphs]
\label{def:botgnngraph}
Given a data instance $e \in {\cal E}$ and $B,\Mu,d$ as before,
we define $BotGNNGraph_{B,\Mu,d}(e)$, or simply
$BotGNNGraph(e)$ =
$TransformGraph(BotGraph_{B,\Mu,d}(e))$
\end{mydefinition}

\begin{figure}[!htb]
    \centering
    \includegraphics[width=0.98\textwidth]{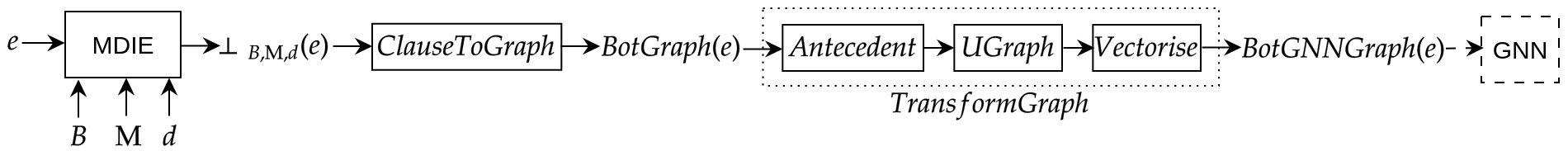}
    \caption{Construction and use of bottom-graphs for use by GNNs in this
    paper. We note that constituting the transformation of bottom-graphs
    are for the GNN implementations used in this paper.}
    \label{fig:botgraphconstr}
\end{figure}

\noindent
Figure ~\ref{fig:botgraphconstr} summarises the sequence
of computations used in this paper.
We will use the term $BotGNN$ to describe GNNs constructed
from BotGNN graphs.





Procedures \ref{algo:train}, \ref{algo:test} use
the definitions we have introduced to
construct and test $BotGNN$ models.\footnote{
In practice, Step \ref{step:train} of Procedure
\ref{algo:train} and Step \ref{step:test} of
Procedure \ref{algo:test} involve
some pre-processing that converts the information
in BotGNN graphs into a syntactic form suitable for
the implementations used. We do not describe these
pre-processing details here: they are available
as code accompanying the paper.}
The procedures assume that data provided as graphs
can be represented as definite clauses (Steps \ref{step:botgraphtrain} in Procedure
\ref{algo:train} and \ref{step:botgraphtest} in Procedure
\ref{algo:test}). We illustrate this with an example.

\begin{myexample}
The chemical Tacrine is a drug used in the treatment of
Alzheimer's disease. It's molecular formula is
$C_{13}H_{14}N_2$, and its molecular structure is shown
in diagrammatic form below:

\begin{center}
    \includegraphics[height=1.5cm]{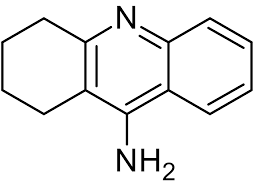}
\end{center}

\noindent
One representation of this molecular graph as a definite
clause is:

\tab $graph(tacrine) \leftarrow$ \\
\tab[2cm] $atom(tacrine,a_1,c)$,\\
\tab[2cm] $atom(tacrine,a_2,c)$,\\
\tab[2cm] \vdots \\
\tab[2cm] $atom(tacrine,a_{13},c)$,\\
\tab[2cm] $atom(tacrine,a_{14},n)$,\\
\tab[2cm] \vdots \\
\tab[2cm] $bond(tacrine,a_1,a_2,1)$,\\
\tab[2cm] $bond(tacrine,a_2,a_3,2)$, \\
\tab[2cm] \vdots
\bigskip

More generally, a graph $g=(V,E,\psi,\phi)$
(where $V$ denotes the vertices, $E$ denotes the edges, $\psi$ and
$\phi$ are vertex and edge-label mappings) can
be transformed into definite clause
of the form $graph(g) \leftarrow Body$, where $Body$ is
a conjunction of ground-literals of the form
$vertex(g,v_1)$, $vertex(g,v_2)$, \ldots;
$edge(g,e_1)$, $edge(g,e_2)$, \ldots;
$vlabel(g,v_1,\psi(v_1))$, $vlabel(g,v_2,\psi(v_2))$, \ldots; and
$elabel(g,e_1,\psi(e_1))$,
$elabel(g,e_2,\psi(e_2))$, \ldots and so on where
$V = \{v_1,v_2,\ldots\}$, $E = \{e_1,e_2,\ldots\}$.
More compact representations are possible, but in the experimental section following, we will be using this kind of simple  transformation (for molecules: the transformation is done automatically from a standard
molecular representation).
\label{ex:tacrine}
\end{myexample}

\begin{algorithm}[!htb]
\SetAlgoLined
\LinesNumbered
\KwData{Background knowledge $B$, modes $\Mu$, depth-limit
    $d$, training data $D_{tr} = \{(g_i,y_i)\}_1^N$, and some
    procedure $TrainGNN$ that trains a graph-based neural
    network}
\KwResult{A $BotGNN$}
\begin{enumerate}[1.]
\item $D'_{tr}$ = $\{$ $(g'_i,y_i): (g_i,y_i) \in D_{tr}$,
    $e_i$ be a ground definite-clause representing $g_i$,
        $g'_i = BotGNNGraph_{B,\Mu,d}(e_i)$ $\}$ \label{step:botgraphtrain}
\item Let $BotGNN = TrainGNN(D'_{tr})$ \label{step:train}
\item \Return $BotGNN$
\end{enumerate}
\caption{\textbf{(TrainBotGNN)} Construct a $BotGNN$ model, given
    training data $\{(g_i,y_i)\}^N_1$, where each $g_i$ is a graph
    and $y_i$ is the class-label for $g_i$.}
\label{algo:train}
\end{algorithm}
    
\begin{algorithm}[!htb]
\SetAlgoLined
\LinesNumbered
\KwData{A $BotGNN$ model, background knowledge $B$, modes $\Mu$, depth-limit
    $d$, and data $D$ consisting of a set of graphs $\{g_i\}_1^N$}
\KwResult{$\{(g_i,\hat{y_i})\}_1^N$ where the $\hat{y_i}$ are predictions
    by $BotGNN$}
\begin{enumerate}[1.]
    \item Let $D'$ = $\{(g_i,g_i'): $
            $g_i \in D$,
            $e_i$ is the definite-clause
            representation of $g_i$,
            $g'_i = BotGNNGraph_{B,\Mu,d}(e_i) \}$\label{step:botgraphtest}
    \item Let $Pred$ = $\{(g_i,\hat{y_i}): (g_i,g_i') \in D', \hat{y_i} = BotGNN(g')\}$ 
    \label{step:test}
    \item \Return $Pred$
\end{enumerate}
\caption{\textbf{(TestBotGNN)} Obtain predictions of a $BotGNN$ model on
    a data set}
\label{algo:test}
\end{algorithm}

\subsection{Note on Differences to Vertex-Enrichment}
\label{sec:note}

While we defer most related work to a later section (Sec.~\ref{sec:relworks}), it
is useful to clarify here some differences of BotGNNs 
with the approach of vertex-enrichment in GNNs
(or VEGNNs). These were introduced in~\citep{dash2021incorporating} with the same goal
of incorporating symbolic domain-knowledge into GNNs. An immediate
difference is in the nature of the graphs handled by the two approaches.
Broadly, VEGNNs require data in a graphical form. VEGNNs
retain the most of the original graph-structure, but modify the
feature-vectors associated with each vertex of the graph (more on this below).
BotGNNs on the other hand do not require data to be a graph. Instead, any
data representable as a definite clause are reformulated using the
bottom-clause into BotGNN graphs. Recall these are
bipartite-graphs, in which both vertices and their labels have
a different meaning to the graphs in VEGNNs.

A subtler difference between BotGNNs and VEGNNs arises from how the relational
information is included within the graphs constructed in each case. The
difference is best illustrated by example.

\begin{myexample}
\label{ex:napthalene}

Suppose we consider a molecule containing the atoms and bonds shown
on the left below, and we want to include the $6$-ary relation of a benzene ring
(the corresponding hyper-edge is shown dotted on the right below). 

\begin{center}
    \includegraphics[height=2cm]{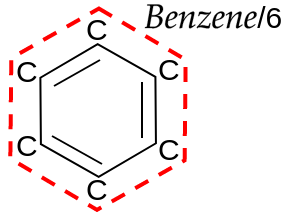}
\end{center}

In VEGNNs~\citep{dash2021incorporating}, graphs are represented  as
tuples of the form $(V,E,\sigma,\psi,\phi)$, where
$V$ is the set of vertices (here atoms in the molecule);
$E$ denotes the edges (bonds in the molecule);
$\sigma$ is a neighbourhood function;
$\psi$ denotes an initial
vertex-labelling; and $\phi$ denotes an initial edge-labelling. 
For each $v \in V$, let $\psi(v)$ be a real-valued vector of finite dimension.
In \citep{dash2021incorporating}, any $n$-ary relation in domain-knowledge 
is treated as a hyperedge, where a hyperedge is a set of $n$ vertices in the graph. 
For any vertex $v$ in a graph, let $h(v)$ denote the set of predicate symbols
such that the corresponding hyper-edge contains $v$. Let
$g/1$ be a function that maps sets of predicate-symbols
to a fixed-length Boolean-valued vector (a ``multi-hot'' encoding).
Thus, in \citep{dash2021incorporating}, 
a VEGNN is a GNN that operates on graphs obtained
from labelled graphs of the form
$(V,E,\sigma,\psi_{V},\phi)$, where $\psi_{V}(v) = \psi(v) \oplus g(h(v))$ (here
$\oplus$ denotes a concatenation operation). In a VEGNN $h(v)$ is $\{Benzene/6\}$
for $v = v_1,\ldots,v_{10}$ in the graph below (representing
the compound naphthalene):

\begin{center}
    \includegraphics[height=2cm]{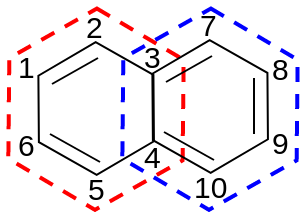}
\end{center}

\noindent
Thus, the information
that $v_3, v_4$ are members of 2 different benzene rings is not captured
in the VEGNNs vertex-labelleing,
and we have to rely on the GNN machinery
to re-derive this information from the graph structure (if this information is needed).
In a BotGNN on the other hand, the two
benzene rings are separate vertices in the bipartite graph,
which share edges to vertices representing $v_3$ and $v_4$.
The broad structure of the VEGNN (only vertex-labels are shown
for clarity) and the BotGNN graphs for naphthalene are shown
below in (a) and (b) respectively:

\begin{center}
    \includegraphics[width=0.8\textwidth]{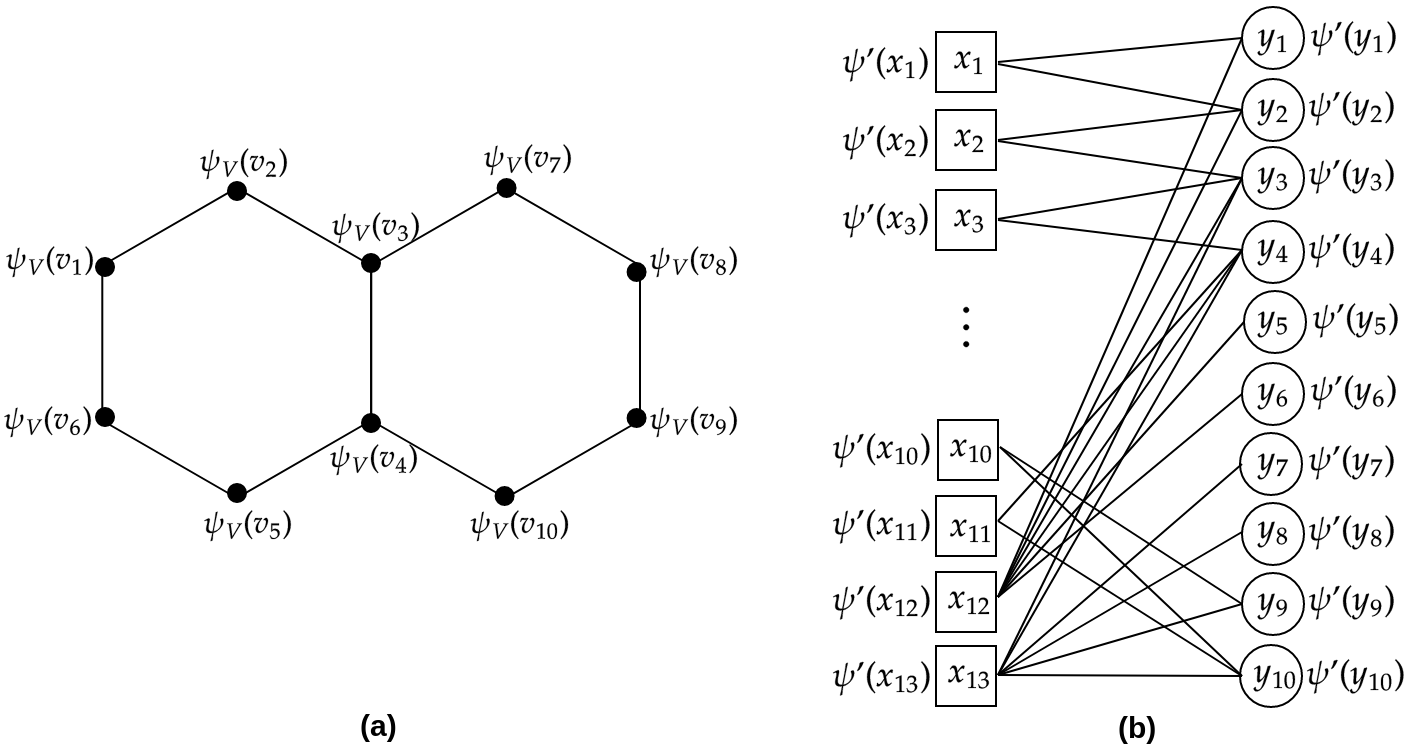}
\end{center}
\noindent
$\psi_V/1$ in (a) refers to the vertex-encoding function in
\citep{dash2021incorporating}, and $\psi'$ in (b) refers to the
function defined in Defn. \ref{def:vectorise}. For the experimental
data in this paper, the vertex-encoding in \citep{dash2021incorporating} results
in vectors whose dimensions are about 10 times more than the $\psi_V/1$ from
Defn. \ref{def:vectorise}.


\end{myexample}


The approach to $n$-ary relations employed by VEGNNs is thus somewhat akin to a
{\em clique-expansion\/} of the graph containing vertices for terms. In a clique-expansion, all
vertices in a hyper-edge--elements of some $n$-ary relation--are connected
together by a labelled hyper-edge. This can introduce a lot of new edges,
and some mechanism is needed to distinguish between multiple occurrences of the same
relation (an example is the multiple occurrences of benzene rings above).
VEGNNs can be seen as achieving the effect of such a clique-expansion,
without explicitly adding the new edges, but they do not
address the problem of multiple occurrences. BotGNNs can be seen instead as a
{\em star-expansion\/} of the graph containing vertices for terms.
In such a star-expansion, new nodes denoting
the relation are introduced, along with edges between the relation-vertex
and the term-vertices that are part of the relation (that is, the
hyper-edge). Star-expansions of graphs thus contain 2 kinds of vertices,
which is similar to the graph constructed by a BotGNN.



\section{Empirical Evaluation}
\label{sec:expt}

\subsection{Aims}
\label{sec:aims}

Our aim in this section is to investigate the following claims:
\begin{enumerate}
    \item GNNs based on bottom-graphs constructed
    with domain knowledge ($BotGNN$s)
    have a higher predictive performance than GNNs that do not
    use domain knowledge;
    \item $BotGNN$s have a higher predictive performance than
    vertex-enriched GNNs ($VEGNN$s) that employ a simplification
    to provide domain knowledge to GNNs~\citep{dash2021incorporating}.
\end{enumerate}

\subsection{Materials}
\label{sec:mat}

\subsubsection{Data}

For the empirical evaluation of our proposed BotGNNs, we
use 73 benchmark datasets arising in the field of drug-discovery.
Each dataset represents extensive drug evaluation effort at the NCI\footnote{The National Cancer Institute (\url{https://www.cancer.gov/})}
to experimentally determine the effectiveness of anti-cancer activity of a compound against a number of cell lines~\citep{marx2003data}. The datasets correspond to the concentration parameter GI50, which is the concentration that results in 50\% growth inhibition. 
Each dataset consists of a set of chemical compounds, which
are then converted into bottom-graphs. 

Each bottom-graph can be represented using $(G,\cdot)$,
where $G = (X, Y, E)$, where $X$ represents the vertices
corresponding to the relations, $Y$ represents the vertices
corresponding to ground terms in the bottom-clause
constructed by MDIE, 
and $E$ represents the edges
between $X$ and $Y$. Table~\ref{tab:dataset} summarises
the datasets.

\begin{table}[!htb]
    \centering
    \begin{tabular}{ccccc}
    \toprule
    \# of  & Avg \# of  & avg. of & avg. of & avg. of\\
    datasets    & instances & $|X|$ & $|Y|$ & $|E|$ \\
    \midrule
    73  &   3032 & 81 & 42 & 937\\
    \bottomrule
    \end{tabular}
    \caption{Dataset summary (The last 3 columns
    are the average number of $X$, $Y$ and $E$ in 
    each bottom-graph in a dataset)}
    \label{tab:dataset}
\end{table}

\subsubsection{Background Knowledge}
The initial version of the background knowledge was used in~\citep{Craenenbroeck2002dmax,ando2006discovering}. 
This BK is a collection of logic
programs (written in Prolog) defining almost $100$
relations for various functional groups (such as amide, amine,
ether, etc.) and various ring structures (such as aromatic, non-aromatic etc.). A functional group
is represented by $\mathtt{functional\_group}/4$ predicate and a ring
is represented by $\mathtt{ring}/4$.
There are also higher-level relations defined on the top
of the above two relations.
These are: the presence of fused rings, connected rings and substructures. 
\begin{description}
    \item[$\mathtt{has\_struc(CompoundId,Atoms,Length,Struc)}$] This relation is $TRUE$ if a compound identified by $\mathtt{CompoundId}$
    contains a structure $\mathtt{Struc}$ of length $\mathtt{Length}$ containing a set of atoms in $\mathtt{Atoms}$.
    \item[$\mathtt{fused(CompoundId,Struc1,Struc2)}$] This relation is $TRUE$ if a compound identified by $\mathtt{CompoundId}$
    contains a pair of fused structures $\mathtt{Struc1}$ and $\mathtt{Struc2}$ (that is, there is
    at least 1 pair of common atoms).
    \item[$\mathtt{connected(CompoundId,Struc1,Struc2)}$] This relation is $TRUE$ if a compound identified by 
    $\mathtt{CompoundId}$
    contains a pair structures $\mathtt{Struc1}$ and $\mathtt{Struc2}$ that are not fused
    but connected by a bond between an atom in $\mathtt{Struc1}$ and an atom in
    $\mathtt{Struc2}$.
\end{description}

\subsubsection{Algorithms and Machines}

The datasets and the BK are written in Prolog. We
use Inductive Logic Programming (ILP) engine, Aleph~\citep{srinivasan2001aleph} to construct
the bottom-clause using MDIE. A Prolog program
then extracts the relations and ground terms
from the bottom-clause. We use YAP compiler
for execution of all our Prolog programs. 
These are parsed by UNIX
and MATLAB scripts to construct bottom-graph datasets
in the format prescribed in \citep{KKMMN2016},
which are mainly representations of adjacency matrix,
vertex labels (feature vector), class labels, etc.

The GNN variants used here are described in the next section.
All the experiments are conducted in a Python environment.
The GNN models have been implemented by using the PyTorch Geometric library~\citep{fey2019fast}--a popular geometric deep learning extension for PyTorch~\citep{paszke2019pytorch} enabling easier implementations of various graph convolution
and pooling methods.

For all the experiments, we use a machine with Ubuntu (16.04 LTS) operating system, and hardware configuration such as:
64GB of main memory, 16-core Intel Xeon processor, a NVIDIA P4000 graphics processor with 8GB of video memory.

\subsection{Method}
\label{sec:meth}

Let $D$ be a set of data-instances represented
as graphs $\{(g_1,y_1), \ldots, (g_N,y_N)\}$,
where $y_i$ is a class label associated with the graph $g_i$.
We also assume  that we have access to
background-knowledge $B$, a set of modes $\Mu$, a depth-limit $d$.
Our method for investigating the performance of $BotGNN$s uses
is straightforward:
\begin{enumerate}[(1)]
    \item Randomly split $D$ into $D_{Tr}$ and $D_{Te}$;
    \item Let $BotGNN$ be the model from Procedure~\ref{algo:train} (TrainBotGNN) with
    backround knowledge $B$, modes $\Mu$, depth-limit $d$, training
    data $D_{Tr}$ and some
    GNN implementation (see below);
    \item Let $GNN$ be the model from the GNN implementation without
    background knowledge, and with $D_{Tr}$;
    \item Let $VEGNN$ be the model using the GNN implementation with
    vertex-enrichment using the background knowledge $B$, and with
    $D_{Tr}$;
    \item Let $D'_{Te}$ = $\{ g_i: (g_i,y_i) \in D_{Te}\}$
    \item Obtain the predictions for $D'_{Te}$  of $BotGNN$ using Procedure~\ref{algo:test} (TestBotGNN) with
    background knowledge $B$, modes $\Mu$, and depth-limit $d$;
    \item Obtain the predictions for $D'_{Te}$ using $GNN$ and $VEGNN$; and
    \item Compare the performance of $BotGNN$, $GNN$ and $VEGNN$.
\end{enumerate}

\begin{figure}[!htb]
    \centering
    \includegraphics[width=0.95\textwidth]{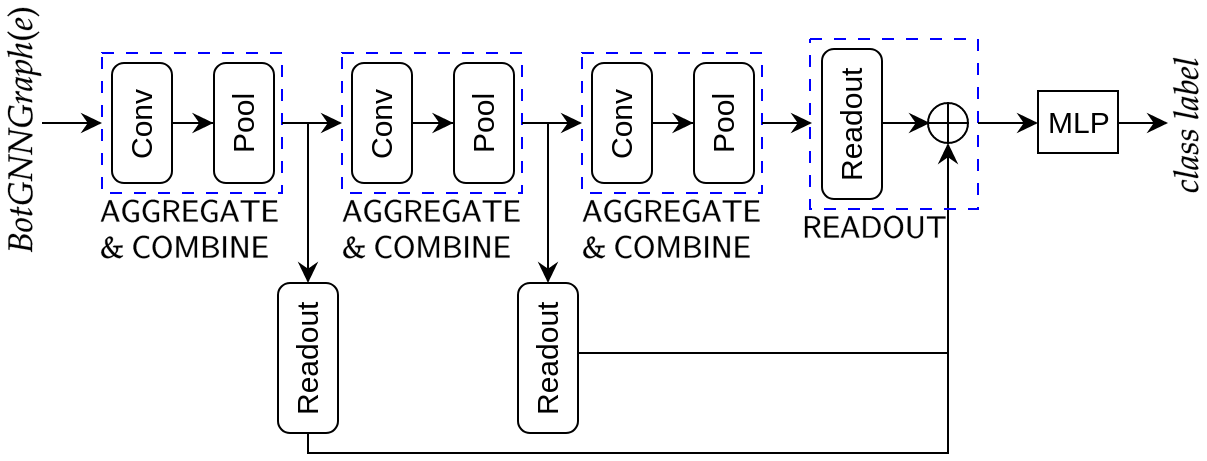}
    \caption{Components involved in implementing the
        workflow in Sec. \ref{sec:gnn}
        for BotGNN models. `Conv' and `Pool' refer
        to the graph-convolution and graph-pooling
        operations, respectively. The `Readout' operation constructs a graph-representation by accumulating
        information from all the vertex in the graph
        obtained after the pooling operation. The final
        graph-representation is obtained in the $\mathsf{READOUT}$ block by an element-wise sum (shown as $\oplus$) of the 
        individual graph-representations obtained
        after each $\mathsf{AGGREGATE}$-$\mathsf{COMBINE}$ block. MLP stands for Multilayer Perceptron.}
    \label{fig:botgnn_arch}
\end{figure}

The following additional details are relevant. 
We closely follow the method used in~\citep{dash2021incorporating} for the construction of GNNs.
The general workflow involved in GNNs was described in Sec.~\ref{sec:gnn}.
A diagram of the components involved in
implementing that workflow
is shown in Fig.~\ref{fig:botgnn_arch}.

\begin{itemize}
    \item We have used a 70:30 train-test split for
    each of the datasets. 10\% of the train-set is 
    used as a validation set for hyperparameter tuning.
    \item Each GNN consists of three graph convolution blocks
    and three graph pooling blocks. The convolution and pooling
    blocks interleave each other (that is, C-P-C-P-C-P) as shown in Fig.~\ref{fig:botgnn_arch}.
    \item The convolution
    blocks can be of one of the following five variants:
    GCN~\citep{Kipf2017gcn},
    $k$-GNN~\citep{morris2019weisfeiler}, 
    GAT~\citep{velickovic2018graph},
    GraphSAGE~\citep{hamilton2017inductive}, 
    and ARMA~\citep{bianchi2021graph}. The
    mathematical details on these graph convolution
    operations are provided in Appendix~\ref{app:gnnmaths}.
    \item The graph pooling block uses self-attention pooling~\citep{lee2019self} with a pooling ratio of 0.5. 
    We use the graph-convolution formula proposed
    in \citep{Kipf2017gcn} for calculating the
    self-attention scores.
    \item Due to the large  number of experiments (resulting
    from multiple datasets and multiple GNN variants),
    the hyperparameters in the convolution
    blocks are set to the default values within the PyTorch Geometric library.
    \item We use a hierarchical pooling architecture that uses the readout mechanism proposed by Cangea~\textit{et al.}~\citep{cangea2018towards}. The readout block aggregates node features to produce a fixed size intermediate representation for 
    the graph. The final fixed-size representation
    for the graph is
    obtained by element-wise addition of the three readout representations.
    \item The representation length ($2m$) is determined by
    using a validation-based approach. The parameter grid for $m$ is: $\{8, 128\}$, representing a small and a large
    embedding, respectively.
    \item The final representation is then fed as input to a 3-layered MLP. We use a dropout layer with a fixed dropout rate of 0.5 after the first layer of MLP.
    \item The input layer of the MLP contains $2m$ units, followed by two hidden layers with $m$ units and $\left \lfloor{m/2}\right \rfloor$ units, respectively. The activation function used in the hidden layers is $\mathtt{relu}$. The output layer uses $\mathtt{logsoftmax}$ activation.
    \item The loss function used is the negative log-likelihood between the target class-labels and the predictions 
    from the model.
    \item We denote the $BotGNN$ variants as: $BotGNN_{1,\ldots,5}$ based on the type of graph convolution method used.
    \item We use the Adam~\citep{kingma2014adam} optimiser for training the BotGNNs ($BotGNN_{1,\ldots,5}$). The learning rate is $0.0005$, weight decay parameter is $0.0001$, the momentum factors are set to the default values of $\beta_{1,2}=(0.9,0.999)$.
    \item The maximum number of training epochs is 1000. 
    The batch size is 128. 
    \item We use an early-stopping mechanism~\citep{prechelt1998early} to avoid
    overfitting during training. The resulting model
    is then saved and can be used for evaluation on 
    the independent test-set. The patience period for 
    early-stopping is fixed at $50$.
    \item The predictive performance of a BotGNN model refers
    to its predictive accuracy on the independent test-set.
    \item Comparison of the predictive performance of BotGNNs
    against GNNs and VEGNNs is conducted using the Wilcoxon
        signed-rank test, using the standard implementation
        within MATLAB (R2018b).
\end{itemize}

\subsection{Results}
\label{sec:results}

\begin{figure}[!htb]
\centering
\begin{subfigure}{.5\textwidth}
  \centering
  \includegraphics[width=\textwidth]{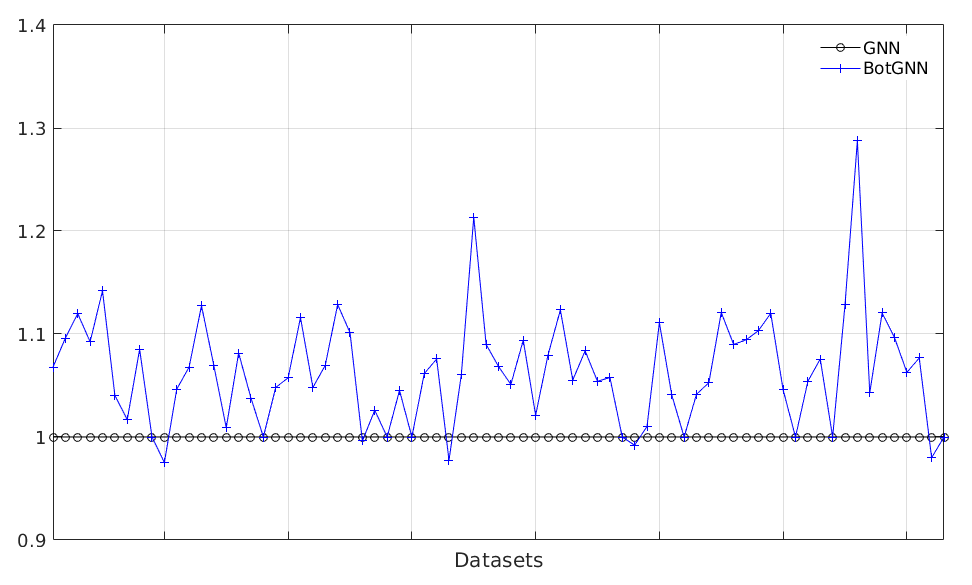}
  \caption{${GNN}_1$ (median gain $\approx 6\%$)}
\end{subfigure}%
\begin{subfigure}{.5\textwidth}
  \centering
  \includegraphics[width=\textwidth]{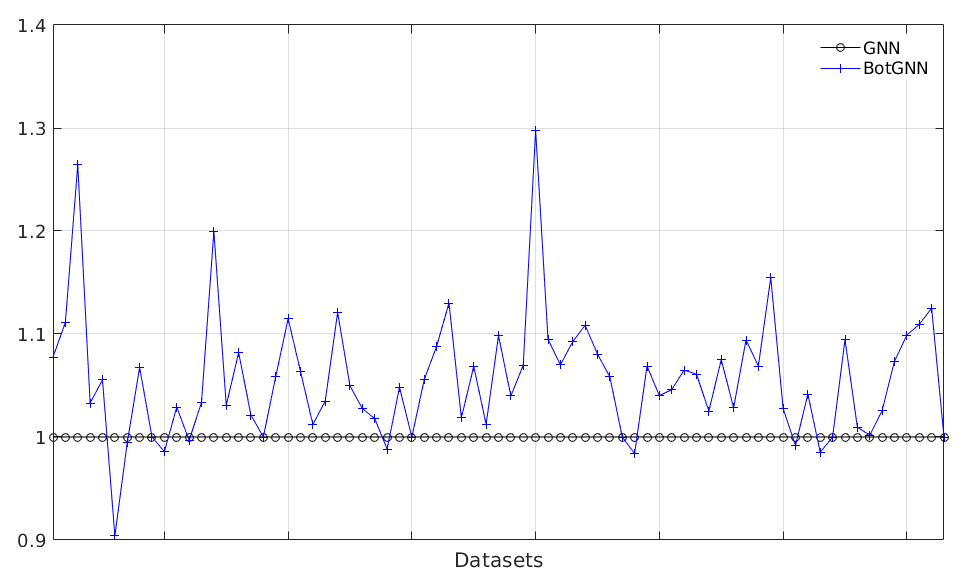}
  \caption{${GNN}_2$ (median gain $\approx 5\%$)}
\end{subfigure}
\begin{subfigure}{.5\textwidth}
  \centering
  \includegraphics[width=\textwidth]{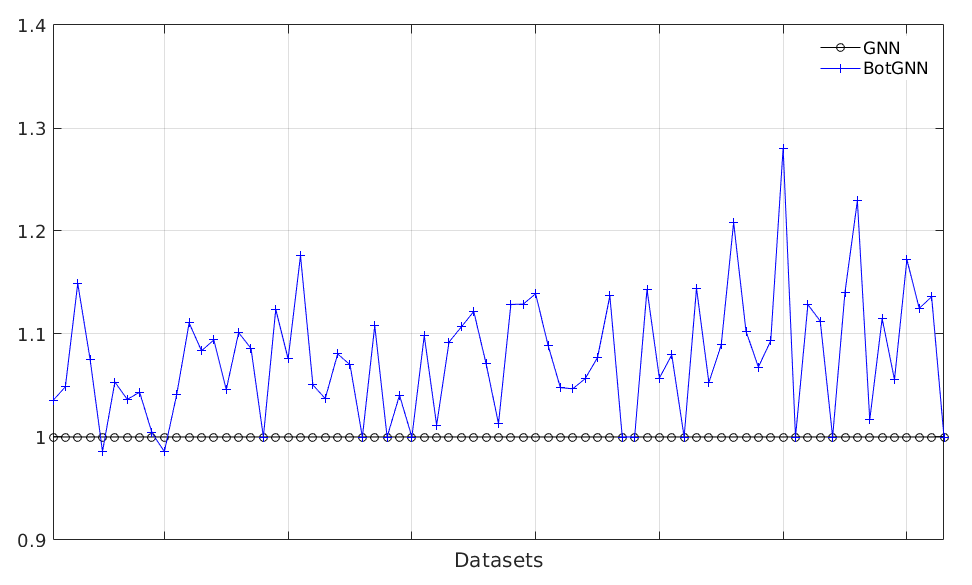}
  \caption{${GNN}_3$ (median gain $\approx 8\%$)}
\end{subfigure}%
\begin{subfigure}{.5\textwidth}
  \centering
  \includegraphics[width=\textwidth]{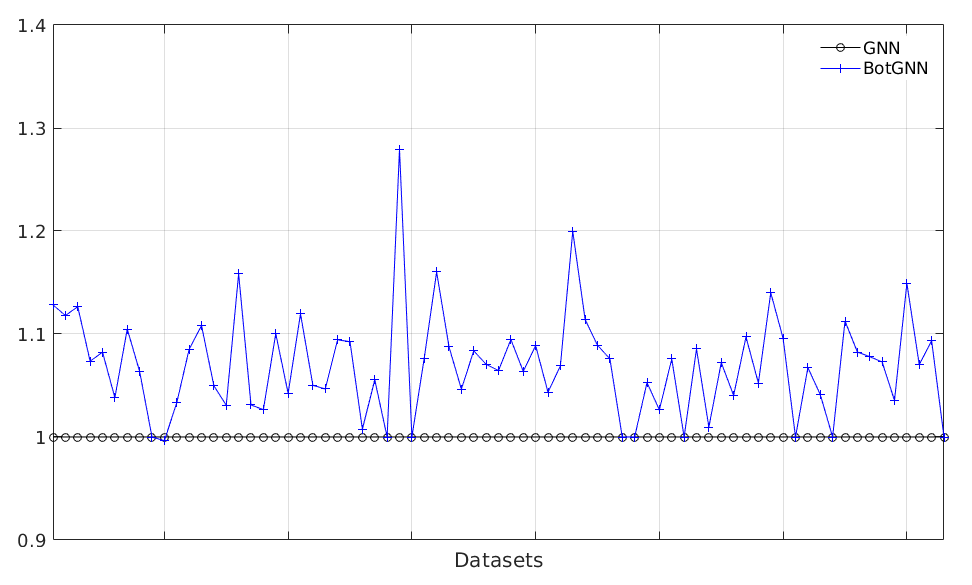}
  \caption{${GNN}_4$ (median gain $\approx 7\%$)}
\end{subfigure}
\begin{subfigure}{.5\textwidth}
  \centering
  \includegraphics[width=\textwidth]{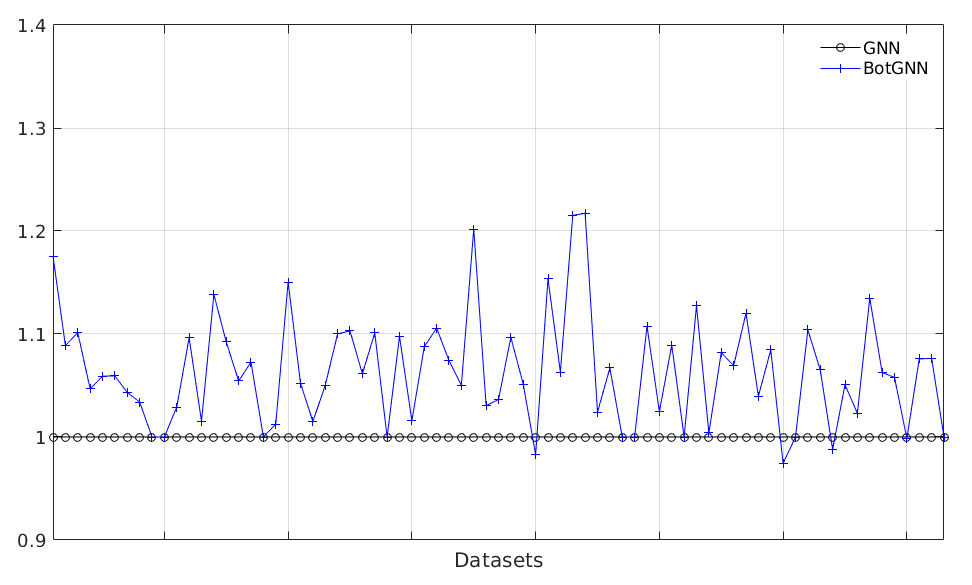}
  \caption{${GNN}_5$ (median gain $\approx 6\%$)}
\end{subfigure}
\caption{Qualitative comparison of predictive performance of
BotGNNs against Baseline (that is, GNN variants without access to domain-relations). Performance refers to estimates
of predictive accuracy (obtained on a holdout set),
and all performances are normalised against that of
baseline performance (taken as $1$).
No significance should be
attached to the line joining the data points: this is only for visual clarity.}
\label{fig:plotsbotgnnvsgnn}
\end{figure}

The quantitative comparisons of predictive performance of $BotGNN$s
against baseline $GNN$s and $VEGNN$s are presented in
Fig.~\ref{fig:results1}. The tabulation shows number of datasets
on which $BotGNN$ has higher, lower or equal predictive accuracy.
The principal conclusions from these tabulations are these:
(1) $BotGNN$s perform significantly better than their
    corresponding counterparts that do not have access
    to any information other than the atom-and-bond
    structure of a molecule achieving a gain in
    predictive accuracy of 5-8\% across variants
    as shown in the qualitative comparison shown
    in Fig.~\ref{fig:plotsbotgnnvsgnn}.
    This is irrespective of the
    variant of GNN used, suggesting that the
    technique is able to usefully integrate domain knowledge;
(2) $BotGNN$s perform significantly better than $VEGNN$s with
    access to the same background knowledge. This suggests that
    $BotGNN$s do more than the vertex-enrichment approach used
    by $VEGNN$s.
    
\begin{figure}[!htb]
\centering
\begin{tabular}{cll}
\toprule
 GNN & \multicolumn{2}{c}{Accuracy ($BotGNN$)} \\
Variant & \multicolumn{2}{c}{Higher/Lower/Equal ($p$-value)} \\ \cline{2-3}
 & \multicolumn{1}{c}{$GNN$} & \multicolumn{1}{c}{$VEGNN$}  \\ 
 \midrule
1 & 59/5/9 ($<$ 0.001) & 54/11/8 ($<$ 0.001) \\ 
2 & 59/8/6 ($<$ 0.001) & 61/9/3 ($<$ 0.001) \\ 
3 & 61/2/10 ($<$ 0.001) & 54/10/9 ($<$ 0.001) \\ 
4 & 63/1/9 ($<$ 0.001) & 55/11/7 ($<$ 0.001) \\ 
5 & 60/4/9 ($<$ 0.001) & 52/9/12 ($<$ 0.001) \\
\bottomrule
\end{tabular}
\caption{Comparison of predictive performance of $BotGNN$s. 
The tabulations are the number of datasets on which $BotGNN$ 
has higher, lower or equal predictive accuracy (obtained on a holdout set)
than $GNN$ and $VEGNN$. 
Statistical significance is computed by the Wilcoxon signed-rank test.}
\label{fig:results1}
\end{figure}
\vspace{0.3cm}

In a previous section (Sec. \ref{sec:note}) we have
described differences between BotGNNs and VEGNNs
arising from an encoding of the data into a bipartite graph
representation. Possible reasons for this difference in
performance are twofold:
(1) The GNN variants are unable
to use edge-label information. In the VEGNN-style
graphs for the data, this information corresponds to the type of bonds.
However, this information is contained in vertices associated with the bond-literals
in BotGNN-style graphs, which can be used by the GNN-variants; and
(2) The potential loss in relational information in VEGNN-style graphs
as described in Sec. \ref{sec:note}.
A further difference, not apparent
from tabulations of performance is the differences in
the feature-vectors associated with each vertex.
For the data here, vertex-labels for VEGNNs
described in \citep{dash2021incorporating}
results in each vertex being associated with a 1400-dimensional
vector. For BotGNNs, this is about 130.

\begin{figure}[!htb]
    \centering
    \begin{minipage}{0.5\textwidth}
    \centering
    \includegraphics[width=0.95\textwidth]{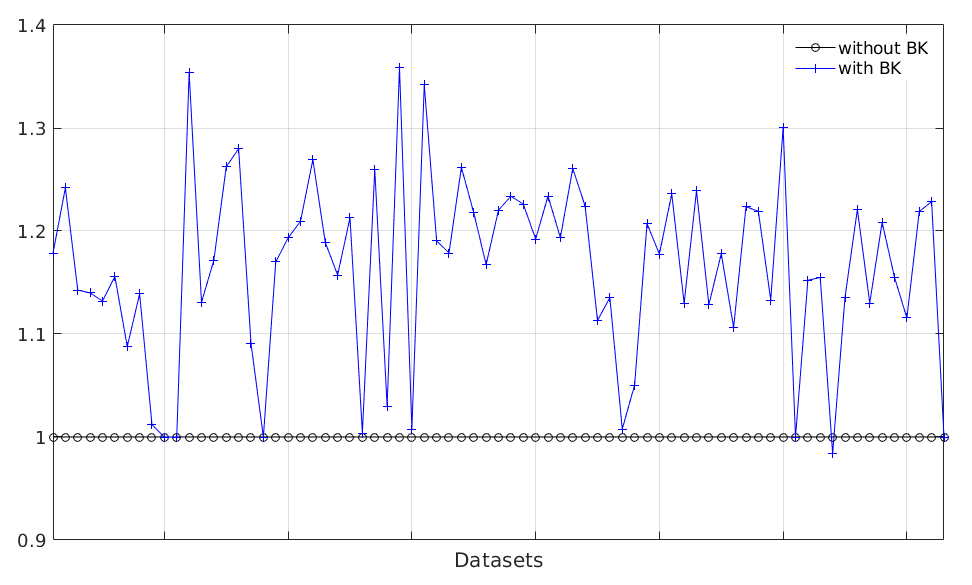}
    \caption*{(a) DRM}
    \end{minipage}%
    \begin{minipage}{0.5\textwidth}
    \centering
    \includegraphics[width=0.95\textwidth]{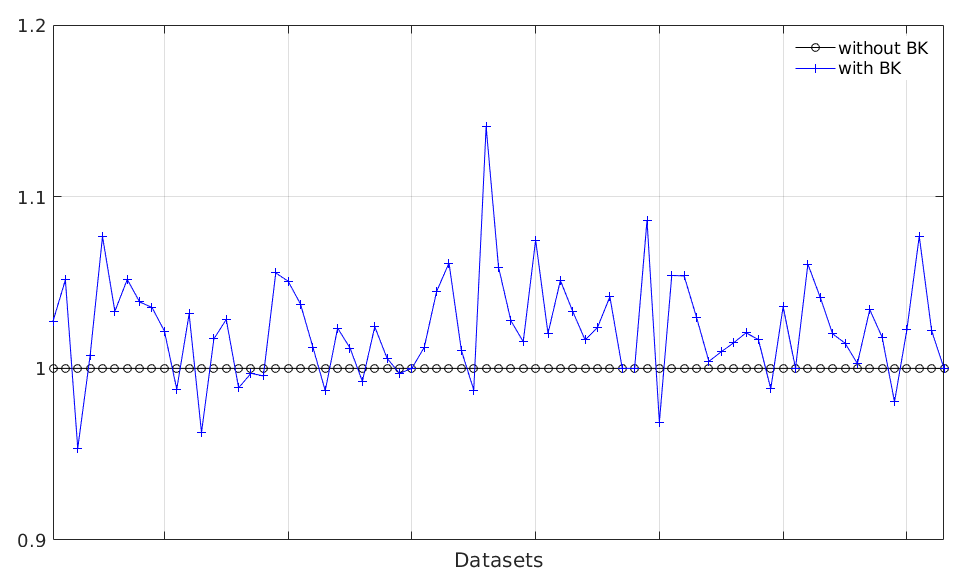}
    \caption*{(b) BCP+MLP}
    \end{minipage}
    \caption{Improvements in predictive performance of
        MLPs, when provided with domain-knowledge 
        through propositionalisation. Baselines (``1'') are the models without domain-knowledge. The
        DRM here is an MLP that
        uses simple Boolean propositions
        indicating the presence or absence of relations
        provided as domain-knowledge.
        The structure and parameters of the MLP are obtained using the Adam optimiser \citep{kingma2014adam}.
        BCP+MLP use Boolean propositions
        constructed using the bottom-clause propositionalisation
        method in \citep{francca2014fast}. ``MLP'' refers to
        the use of these features by a multi-layer perceptron.
        The structure and parameters for the MLP are obtained
        using the same approach as the DRM. The
        domain-knowledge is the same as that used
        for the construction of BotGNNs.}
    \label{fig:propcompare1}
\end{figure}

The significant improvement in predictive performance
with the inclusion of symbolic domain-knowledge
into GNNs is consistent
with similar observations we obtain with
multilayer perceptrons (MLPs). For the
latter, one well-established way of inclusion
of domain-knowledge is through the use of
the technique of {\em propositionalisation}
\citep{lavravc1991learning}. This represents
relational data, like graphs, in the form of 
some numeric vector (usually, Boolean).
For instance, a simple and effective method known as 
``bottom-clause propositionalisation'' or BCP
\citep{francca2014fast} is a propositionalisation
approach serving as an extension to one of the 
pioneering works on neural-symbolic learning systems
proposed in \citep{garcez1999connectionist}.
For a set of (relational) data-instances,
BCP obtains a set of (unique) literals from the 
most-specific (or bottom) clauses constructed by an ILP engine.
It then propositionalises each bottom-clause based on
the literal set. 
This process results in each relational data-instance 
being represented as a Boolean feature-vector.
These feature-vectors are then input to an MLP
for further processing, allowing the symbolic domain-information
available in the bottom-clauses to be easily
incorporated into the MLP.
Some other related studies have shown that
the most-specific clauses can also be treated as a source of
relational features, 
described by conjunctions of literals.
These relational features
can be used to construct a Boolean-vector representation of
the (relational) data-instances. 
These features can form the input feature-vectors
for standard statistical models (as in \citep{saha2012kinds})
or for multilayer perceptrons (MLPs).
When used with MLPs, the resulting model is called as
``deep relational machines'' or DRMs, as introduced by \citep{lodhi2013deep} and studied extensively in
\citep{dash2018large,dash2019discrete}.
In Fig.~\ref{fig:propcompare1} we also observe
gains in performance of MLPs by incorporating
domain-knowledge through the use of some
form of propositionalisation.

\subsection{Some Additional Results}
\label{sec:additional}

We turn now to two questions that are not
within the scope of the experimental goals
listed in Sec. \ref{sec:aims}, but are nevertheless
of practical interest (details relevant to the results
are in Appendix \ref{app:appdetails}). First, the question of
whether incorporating domain-knowledge using
the route of bottom-graphs and GNNs is better
than propositions and MLPs?
A straightforward comparison
of the $BotGNN$ models against those used in Fig.~\ref{fig:propcompare1}
would seem to suggest that the answer is ``yes'' (Fig.~\ref{fig:propcompare2}).
However, we caution against drawing such a conclusion
for at least the following reasons:
(a) Inclusion of propositions based on
    stochastic sampling of
    complex relational features
    in  \citep{dash2019discrete} can
    result in significantly better DRM models.
    It is possible also that BCP could be augmented with
    the same sampling methods to yield more
    informative propositions;
    (b) It is also possible that a $BotGNN$
        obtained with access to sampled relational
        features could improve its performance over
        what is shown here.
    We note that propositionalisation is not confined to the
use of MLPs: so it would be not be surprising if the BotGNN
performance was bettered by some ML method using
the data provided to the DRM or BCP+MLP. 

A more useful difference between the BotGNN approach and
    propositionalisation is that techniques relying on the
    latter usually separate the feature- and
    model-construction steps. A $BotGNN$, like any GNN,
    constructs a vector-space embedding for the graphs it
    is provided. However, this embedding is obtained
    as part of an end-to-end model construction
    process. This can be substantially
    more compact than the representation used by methods
    that employ a separate propositionalisation step
    (see Fig.~\ref{fig:propsize}).

\begin{figure}[!htb]
    \centering
    \begin{tabular}{cll}
        \toprule
        GNN & \multicolumn{2}{c}{Accuracy ($BotGNN$)} \\
        Variant & \multicolumn{2}{c}{Higher/Lower/Equal ($p$-value)} \\ \cline{2-3}
        & \multicolumn{1}{c}{DRM} & \multicolumn{1}{c}{BCP+MLP}  \\ 
        \midrule
        1 & 47/22/4 ($<$ 0.001) & 58/10/5 ($<$ 0.001) \\
        2 & 41/29/3 (0.065) & 58/11/4 ($<$ 0.001) \\
        3 & 45/19/9 ($<$ 0.001) & 61/6/6 ($<$ 0.001) \\
        4 & 50/19/4 ($<$ 0.001) & 62/6/5 ($<$ 0.001) \\
        5 & 51/16/6 ($<$ 0.001) & 60/6/7 ($<$ 0.001) \\
        \bottomrule
    \end{tabular}
    \caption{Comparison of predictive performance of BotGNN with DRM and BCP+MLP. 
    The tabulations are the number of datasets on which $BotGNN$ 
    has higher, lower or equal predictive accuracy (obtained on a holdout set)
    than DRM and BCP+MLP. 
    DRM and BCP+MLP  refer to
    the models in Fig.~\ref{fig:propcompare1}}
\label{fig:propcompare2}
\end{figure}

\begin{figure}[!htb]
    \centering
    \begin{tabular}{ccc}
    \toprule
    Method & Vector Representation & Vector Dimension (Range) \\
    \midrule
    BotGNN & Real, dense & 16--256\\
    DRM & Boolean, sparse & 1400--1400 \\ 
    BCP+MLP & Boolean, very sparse & 18000--52000 \\ 
    \bottomrule
    \end{tabular}
    \caption{Characterisation of vector-representation used
    for model-construction by
    BotGNNs, DRMs and BCP+MLP.
    Minimum/maximum values of the range are
    only shown to 3 meaningful digits (the
    actual values are not relevant here).
    The graph-representations (also, called graph-embeddings) 
    for BotGNNs are constructed
    internally by the GNN. By ``sparse'' we mean
    that there are many $0$-values, and by
    ``very sparse'', we mean the values are mostly $0$.
    }
    \label{fig:propsize}
\end{figure}

The second question of practical interest is how
a BotGNN's performance compares to an ILP learner that
uses bottom-clauses directly. Fig.~\ref{fig:ilpcompare}(a) shows a routine comparison against the Aleph system \citep{srinivasan2001aleph}
configured to perform greedy set-covering for
identifying rules using bottom-clauses
(in effect, a Progol-like approach: see \citep{muggleton1995inverse}).
Again, we caution against drawing the obvious
conclusion, since the results are obtained 
without attempting to optimise any parameters of
the ILP learner (only the minimum accuracy of clauses
was changed from the default setting of $1.0$ to $0.7$: this
latter value has been shown to be more appropriate in many
previous experimental studies with Aleph).
A better indication is in Fig.~\ref{fig:ilpcompare}(b),
which compares BotGNN performance on older benchmarks
for which ILP results after parameter optimisation are
available. These suggest that BotGNN
performance to be comparable to an ILP approach
with optimised parameter settings.\footnote{We
note that parameter screening and
optimisation is not routinely
done in ILP. In \citep{srinivasan2011parameter} it is noted:
``Reports in the [ILP] literature rarely contain any discussion of
sensitive parameters of the system or their values. Of 100
experimental studies reported in papers presented between 1998 and 2008
to the principal conference in the area, none attempt any form of
screening for relevant parameters.''}

\begin{figure}[!htb]
    \centering
    \begin{subfigure}{.6\textwidth}
        \centering
        \begin{tabular}{cc}
        \toprule
        GNN & Accuracy ($BotGNN$)\\
        Variant & Higher/Lower/Equal ($p$-value) \\
        \midrule
        1 & 62/7/4 ($< 0.001$) \\
        2 & 60/9/4 ($< 0.001$) \\
        3 & 61/7/5 ($< 0.001$) \\
        4 & 62/6/5 ($< 0.001$) \\
        5 & 62/4/7 ($< 0.001$) \\
        \bottomrule
        \end{tabular}
        \caption*{(a)}
    \end{subfigure}%
    \begin{subfigure}{.4\textwidth}
        \centering
        \begin{tabular}{ccc}
        \toprule
        Dataset & ILP & BotGNN\\
        \midrule
        DssTox & 0.73 & 0.76 \\
        Mutag & 0.88 & 0.89 \\
        Canc & 0.58 & 0.64 \\
        Amine & 0.80 & 0.84 \\
        Choline & 0.77 & 0.72 \\
        Scop & 0.67 & 0.65 \\
        Toxic & 0.87 & 0.85 \\
        \bottomrule
    \end{tabular}
    \caption*{(b)}
    \end{subfigure}
    \caption{Comparison of predictive performance of BotGNNs with an ILP learner (Aleph system): (a) Without hyperparameter tuning in Aleph;
    (b) With hyperparameter tuning. 
    In (a), the tabulations are the number of datasets on which $BotGNN$ 
    has higher, lower or equal predictive accuracy (obtained on a holdout set)
    than the ILP learner. 
    In (b), each 
    entry is the average of the accuracy 
    obtained across 10-fold validation splits 
    (as in \citep{srinivasan2003empirical})}
    \label{fig:ilpcompare}
\end{figure}

\section{Related Work}
\label{sec:relworks}

One of the oldest approaches that incorporate domain knowledge into feature-based
machine learning is LINUS~\citep{lavravc1991learning},
which proposed the idea of an attribute-value
learning system based on a method called
`propositionalisation'. 
This is a simple and effective approach to construct
a numeric feature-vector representation for 
domain-relations~\citep{lavrac2021prop}, 
which can then be used as input features for
deep neural networks. One such example is Deep Relational Machines (DRMs:~\citep{lodhi2013deep}). Some large-scale
studies show that propositionalisation of relational
(first-order) features results in an effective way of incorporating symbolic domain-knowledge into
deep networks~\citep{dash2018large,dash2019discrete}.
The pioneering work on connectionist inductive learning
and logic programming (CIL2P) by \cite{garcez1999connectionist}
is extended in an interesting idea of
propositionalisation 
called `Bottom Clause Proposionalisation (BCP)'~\citep{francca2014fast}. BCP converts a bottom-clauses
in ILP into Boolean vectors,
which can then be used to learn a neural network more efficiently and faster than its predecessor.
Although propositionalisation approaches are a simple and
straight-forward way of constructing a feature-vector
that encodes both relational data and domain-knowledge,
the present forefront of deep networks dealing directly 
with relational data (graphs) are GNNs, 
where we can represent relational
knowledge in neural networks directly without a propositionalisation step.

Other approaches of
incorporating domain-knowledge include a modification
to the loss function that is optimised during training
a deep network~\citep{xu2018semantic,fischer2019dl2}. 
In these approaches, the primary structure
of the underlying deep network stays roughly unaltered.
One such example is: Domain-adapted
neural network (DANN) that introduces an
(additional) domain-based loss term to the neural network loss function. In particular, the domain-specific
rules are approximated using two kinds of constraints:
approximation constraint and monotonicity constraint.
It is claimed that incorporating domain-knowledge
in this manner enforces the network not only to learn
from the available training data but also domain-specific 
rules~\citep{muralidhar2018incorporating}.

Under the category of statistical relational learning (SRL),
there are some interesting proposals to learn from
relational data (often, graph-structured),
and symbolic domain-knowledge. For instance, the work
on kLog by \cite{frasconi2014klog} introduced a method
called `graphicalisation' to convert relational structures
(first-order logic interpretations) to graphs. By this conversion, it enables the use of graph-kernel
methods, which measures the similarity between
two graphs\footnote{Graph Kernel: A kernel function
compares substructures of graphs that are computable in
polynomial time~\citep{vishwanathan2010graph}.}. 
kLog generates propositional features (based on the
graph kernel) for use in SRL.
kLog and BotGNN share some common characteristics: 
(1) at the first level of modelling, they both construct bipartite graphs representing the 
relational instances: kLog constructs undirected bipartite
graphs that are related to 
`Probabilistic Entity-Relationship model' \citep{heckerman2007probabilistic}, 
whereas our approach here constructs directed bipartite graph,
and the construction is different from the former
representation to a great extent, relying heavily on
a vertex-labelling based on the concept of a
mode-language used within ILP; 
(2) at the second level of learning, kLog employs graph
kernels to construct propositional features and uses those
to learn a statistical learner, whereas BotGNN leverages
GNNs, which are superior to graph kernels in practice~\citep{du2019graph}. Though there are some
similarities and differences between kLog and BotGNN,
one should note that the primary aim of the former is 
a proposal for a language for statistical relational
learning, whereas our primary aim is to propose a principled
way to incorporate symbolic domain-knowledge into GNNs.

Falling under the same umbrella of SRL is work on the integration of relational learning 
with GNNs \citep{sourek2021beyond}, which demonstrates
how simple relational logic programs can capture advanced
graph convolution operations in a tightly integrated manner,
requiring the use of a language of Lifted Relational Neural
Networks (LRNNs)~\citep{sourek2018lifted}. The integration
of logic programs and GNNs in this manner results in an interesting GNN-based neuro-symbolic model~\citep{lamb2020graph}.
The input representation for LRNN is a weighted logic program or \textit{template}, which is mainly different from the input representation used for BotGNNs.

Knowledge-graphs (KGs) are a rich source of domain-knowledge and
are a representation of binary relations. In a KG, each vertex
represents an entity. The relation between two vertices
is represented as an edge between them.
In the last few years,
several methods have attempted to incorporate the
information (relations) encoded in KGs into deep neural networks.
For instance, 
\cite{schlichtkrull2018modeling} introduced Relational
Graph Convolutional Networks (R-GCNs) that models a 
information exchange among different entities via a
relation with the help of the message-passing in a GCN.
The approach was intended for entity classification and
link prediction (discovering the missing relation between two
entities). R-GCNs are related to our BotGNNs
in one sense that they are able to model multi-relational
information. However, R-GCNs are restricted to KGs to 
deal only with binary relations, albeit they could be extended
to go beyond binary relations using hypergraph neural 
networks~\citep{feng2019hypergraph,bai2021hypergraph}.
A method proposed for incorporating knowledge-graphs into 
deep networks is termed as ``knowledge-infused learning''
~\citep{kursuncu2019knowledge,sheth2019shades}.
The work examines techniques for incorporating relations at
various layers of deep networks (the authors categorise these as: shallow, semi-deep and deep infusion).
A GNN can directly operate
on knowledge-graphs for constructing 
node and graph representations that are useful
for further learning~\citep{wang2019knowledge}.
We note that BotGNNs could
be seen as performing a generalised
form of knowledge-infused learning, in which: (a) Data can contain
$n$-ary relations; and (b) Domain-knowledge
can encode arbitrary relations of
possible relevance to the data. In
the original work on knowledge-infusion,
data are knowledge-graphs that only
employ binary relations, and there is no
possibility of including additional domain-knowledge
(that is, $B = \emptyset$). 

A method~\citep{xie2019embedding} proposed that the symbolic knowledge can be
represented as formulas in Conjunctive Normal Form
(CNF) and decision-Deterministic Decomposable Negation Normal Form (d-DNNF). These formulas can naturally
be viewed as graph structures. Learning on these
graph structures is then carried out using a GNN. 
A recently proposed method
uses the idea of treating symbolic domain-relations as hyperedges~\citep{dash2021incorporating}.
These hyperedges can then be used to construct the
labelling for the nodes of a graph using a method called `vertex-enrichment', which is a
simplified approach to incorporate symbolic
domain-knowledge into GNNs. This form of GNNs are called
Vertex-Enriched GNNs (VEGNNs).
A detailed note on the differences between VEGNNs and BotGNNs
are already provided in an earlier section.
Briefly, there are three main differences:
(1) VEGNNs require data to be represented as graphs;
whereas, BotGNNs can deal with any data that can be
represented as definite clauses;
(2) VEGNNs introduce symbolic domain-relations 
by modifying only the
vertex-labelling of the graphs while maintaining
the original graph structure of the data; 
whereas, BotGNNs combine data and background knowledge
(using MDIE) to construct BotGNN graph representations,
which are bipartite graphs;
and (3) VEGNNs do not allow some of the crucial information
about a vertex to be automatically conveyed via the
vertex-labellings, for example, a vertex is a member
of two different benzene rings; whereas, in BotGNNs
this information is readily available from the
bipartite graph-structure.
A recent systematic review on various methods of
incorporating domain-knowledge into deep neural networks
categorises domain-knowledge into two classes of constraints: logical and numerical~\citep{dash2021survey}.
Our present work falls under the former category and 
aims to constrain the structure of the graph (here,
bipartite graph). An extended version of this survey
can be found in \citep{dash2021how} which categories
the methods of incorporating domain-knowledge
into deep networks based on whether 
(a) the input-representation is changed,
(b) the loss-function is changed, or
(c) the structure of the deep network is changed.
Our proposed BotGNN approach falls
under the first category where each (relational) data-instance 
is changed to a bipartite graphs representation.

\section{Conclusions}
\label{sec:concl}

The Domain-Knowledge Grand Challenge in \citep{stevens2020ai} 
calls for systematically incorporating
diverse forms of domain knowledge. In this paper,
we have proposed a systematic way of incorporating
domain-knowledge encoded in a powerful subset of
first-order logic.
The technique explicitly addresses the requirement
in \citep{stevens2020ai} of ``extending'' the raw data for
a class of neural networks that deal with graphs.
The significant improvements in performance that we have
observed support the statement on the importance of
the role of domain knowledge. We have also provided
additional results that suggest that the technique
may be doing more than a simple ``propositionalisation''.

On the face of it, it would seem that logic programs
are necessary to construct the BotGNNs proposed here. We distinguish
here between the principle we have proposed and its implementation
using logic programs. The construction of a most-specific
clause, as is done by MDIE, is necessary for
the construction of a BotGNN. However, the construction
of this clause need not be done by using logic programming
technology: it has been shown, for example, in~\citep{bravo2005framework}
how this same formula can be obtained entirely using operations
on relational databases. In many ways, such an implementation
would be more convenient, since the information in
modes can be incorporated (and even augmented) by
the schema of a relational database. Looking further
ahead, it is possible that domain-knowledge could even
be communicated as unstructured statements in a natural
language. However, for problems of scientific discovery
it would appear to be more useful if domain-knowledge was
represented in some structured, mathematical form.

We do not view
BotGNNs as an alternative to ILP. Instead, the
purpose in this paper is to show that techniques developed
in ILP can be used to incorporate symbolic domain
knowledge into deep neural networks, with
significant improvement in predictive performance.
In fact, relational definitions found
by an ILP engine may be able to improve a BotGNN's performance
further, possibly by augmenting bottom-graphs. This can
be done by either simple inclusion of the new relations as
background knowledge, or through an extension of clause-graphs
from bipartite to more general $k$-partite graphs.

The linking of symbolic and neural techniques allows
the possibility of providing logical
explanations for predictions made by the neural model.
This potential has long been recognised, and demonstrated
(see for example:~\citep{besold2017neural,garcez2019neural}). Here, we expect the
relationships shown in Remark~\ref{rem:expl} (see Appendix~\ref{app:ClauseToGraphprop}) may open the
interesting possibility of linking clausal
explanations to GNNs. The improvement in predictive
performance of GNNs by the incorporation of domain knowledge
is a necessary part of their use as tools for scientific
discovery. But that is not sufficient: explanations in
terms of relevant domain-concepts will also be needed.
We intend to look at this in future.

\section*{Data and Code Availability}

Data and codes used in our experiments are available at:
\url{https://github.com/tirtharajdash/BotGNN}.

\begin{acknowledgements}
A.S. is a Visiting Professorial Fellow at School of CSE, UNSW Sydney.
He is also the Class of 1981 Chair Professor at BITS Pilani. 
We thank Gustav {\v{S}}ourek, Czech Technical University, Prague for providing the dataset information; 
and researchers at the DTAI, University of Leuven, for suggestions on how to use the background knowledge within DMAX. 
We also thank Oghenejokpeme I. Orhobor and Ross D. King for providing us with the
initial set of background-knowledge definitions. 
We thank Artur d'Avila Garcez, City, University of London for providing information on the usage of
BCP within their system CILP++.
\end{acknowledgements}

\bibliography{botgnn}

\appendix

\section{Some Properties of Clause-Graphs}
\label{app:ClauseToGraphprop}

We note the following properties about clause-graphs. For these 
properties, we assume
background knowledge $B$,
a set of modes $\Mu$, and a depth-limit $d$ as
before. Clause-graphs are elements of
the set ${\cal G}$ and are structures
of the form $((X,Y,E),\psi)$ where
$(X,Y,E)$ are bipartite graphs from
the set ${\cal B}$ (see Sec. \ref{sec:prelim}).
We assume ${\cal G}$ contains
the element ${CG}_\top$ = 
$((\emptyset,\emptyset,\emptyset),\emptyset)$.
 We also define
the following equality relation
over elements of ${\cal G}$:
$(X_i,Y_i,E_i),\psi_i)$ $=$
$(X_j,Y_j,E_j),\psi_j)$ iff
$X_i = X_j$, $Y_i = Y_j$, 
$E_i = E_j$ and $\psi_i = \psi_j$.
Also, given a clause $C$ =
$\{l_1,\ldots,l_m,\neg l_{m+1},\ldots,\neg l_k\}$,
$1 \leq m < k$,
$\Lambda_C$ is the set
$\{l_1,\ldots,l_{m+1},\ldots,l_k\}$. 


\begin{mydefinition}[$\preceq_{cg}$]
Let ${CG}_1 = (G_1,\psi_1)$, ${CG}_2 = (G_2,\psi_2)$ be
elements of ${\cal G}$, where $G_1 = (X_1,Y_1,E_1)$
and $G_2 = (X_2,Y_2,E_2)$
Then ${CG}_1 \preceq_{cg} {CG}_2$ iff:
(a) $X_1 \subseteq X_2$;
(b) $Y_1 \subseteq Y_2$;
(c) $E_1 \subseteq E_2$; and
(d) $\psi_1 \subseteq \psi_2$.
\label{def:cg}
\end{mydefinition}

\begin{myproposition}
$\langle {\cal {G}}, \preceq_{cg} \rangle$ is partially ordered.
\end{myproposition}

\begin{myproof}
In the following, 
let $CG = ((X,Y,E),\psi)$, and
${CG}_{i} = ((X_i,Y_i,E_i),\psi_i)$.
\begin{description}
\item[Reflexive.] If $CG \in {\cal G}$ then
    $CG \preceq_{cg} CG$. This follows trivially
    since $X \subseteq X$, $Y \subseteq Y$,
    $E \subseteq E$ and $\psi \subseteq \psi$.
\item[Anti-Symmetric.]
Let ${CG}_1, {CG}_2 \in {\cal G}$.
    If
    ${CG}_1 \preceq_{cg} {CG}_2$ and ${CG}_2 \preceq_{cg} {CG}_1$ then
    ${CG}_1 = {CG}_2$.
    Since ${CG}_1 \preceq_{cg} {CG}_2$,
    and ${CG}_2 \preceq_{cg} {CG}_1$
    $X_1 \subseteq X_2$ and $X_2 \subseteq X_1$. Therefore
    $X_1 = X_2$. Similarly
    $Y_1 = Y_2$, $E_1 = E_2$
    and $\psi_1 = \psi_2$,
    and therefore ${CG}_1 = {CG}_2$;
\item[Transitive.] Let ${CG}_1, {CG}_2, {CG}_3 \in {\cal G}$. 
If ${CG}_1 \preceq_{cg} {CG}_2$ and ${CG}_2 \preceq_{cg} {CG}_3$ then
${CG}_1 \preceq_{cg} {CG}_3$.
Since ${CG}_1 \preceq_{cg} {CG}_2$
and ${CG}_2 \preceq_{cg} {CG}_3$ then
$X_1 \subseteq X_2$ and $X_2 \subseteq X_3$. Therefore $X_1 \subseteq X_3$. 
Similarly, $Y_1 \subseteq Y_3$,
$E_1 \subseteq E_3$ and $\psi_1 \subseteq \psi_3$. Therefore, ${CG}_1 \preceq_{cg} {CG}_3$.
\end{description}
\label{prop:cgorder}
\end{myproof}

\smallskip
\noindent
For ${CG}_1, {CG}_2 \in {\cal G}$, if
${CG}_1 \preceq_{cg} {CG}_2$, then we will
say ${CG}_1$ is more general than ${CG}_2$.
We note without formal proof that
if ${CG} \in {\cal G}$ then
    ${CG}_\top \preceq_{cg} CG$.

\begin{myremark}
We note the following consequences of the Defns.
\ref{def:litset}--\ref{def:ClauseToGraph}, and Defn.
\ref{def:cg} above:
\begin{enumerate}[(i)]
    \item Let $C, D \in {\cal L}_{B,\Mu,d}$,
        ${CG}_1 = ClauseToGraph(C)$, and
        ${CG}_2 = ClauseToGraph(D)$ where:
        ${CG}_1 = ((X_1,Y_1,E_1),\psi_1)$ and
        ${CG}_2 = ((X_2,Y_2,E_2),\psi_2)$.
        If $(X_1 \subseteq X_2)$ then
        ${CG}_1 \preceq_{cg} {CG}_2$.
        By construction,
        $X_1 \subseteq X_2$ iff
        $Lits(C) \subseteq Lits(D)$.
        It follows that $Terms(C) \subseteq Terms(D)$,
        and $Y_1 \subseteq Y_2$. Since 
        $E_1$ contains all the relevant arcs
        between $X_1$ and $Y_1$ and
        $E_2$ contains all the relevant arcs
        between $X_2$ and $Y_2$, $E_2$ will
        contain all the elements of $E_1$.
        Since $h_x, h_y$ are bijections,
        $\psi_1 \subseteq \psi_2$. That
        is ${CG}_1 \preceq_{cg} {CG}_2$; \label{rem:xsubs}
    \item Let $C,D \in {\cal L}_{B,\Mu,d}$,
        ${CG}_1 = ClauseToGraph(C) = ((X_1,Y_1,E_1),\psi_1)$ and
        ${CG}_2 = ClauseToGraph(D) = ((X_2,Y_2,E_2),\psi_2)$.
        Let ${\cal {LM}}_1$ be the set of $\lambda\mu$-sequences for
        $C$ and ${\cal {LM}}_2$ be the set of $\lambda\mu$-sequences for $D$. If ${\cal {LM}}_1 \subseteq {\cal {LM}}_2$
        then ${CG}_1 \preceq_{cg} {CG}_2$.
        It is evident that $Lits(C) \subseteq Lits(D)$.
        Therefore $X_1 \subseteq X_2$, and from
        the observation (\ref{rem:xsubs}) above,
        ${CG}_1 \preceq_{cg} {CG}_2)$.\label{rem:lmsubs}
    \end{enumerate}
    \label{rem:subs}
\end{myremark}

\begin{mylemma}[$Lits$]
The function $Lits: {\cal L}_{B,M,d} \rightarrow 2^{LM}$ (defined in Defn. \ref{def:litset}) is well-defined.
That is, if $C = D$, then $Lits(C) = Lits(D)$.
\label{lem:lits}
\end{mylemma}

\begin{myproof}
Assume the contrary. That is,
$C = D$ and $Lits(C) \neq Lits(D)$.
Since $C = D$, $\Lambda_{C} = \Lambda_{D}$.
Further, since $Lits(C) \neq Lits(D)$, for
some $\lambda_i \in \Lambda_{C}, \Lambda_{D}$,
there must exist $\mu_i \in \Mu$ s.t.
$(\lambda_i,\mu_i) \in Lits(C)$
and $(\lambda_i,\mu_i) \not \in Lits(D)$ or {\em vice versa\/}.
This is not possible since $Lits(C)$ and $Lits(D)$ contain all
    $\lambda\mu$-sequences for $C, D$.
\end{myproof}

\begin{mylemma}
Let $C, D \in {\cal L}_{B,\Mu,d}$. 
Let ${CG}_1 = ClauseToGraph(C)$ and
${CG}_2 = ClauseToGraph(D)$. 
if $C = D$ then ${CG}_1 = {CG}_2$.
\label{lem:cg}
\end{mylemma}

\begin{myproof}
The result holds trivially if $C = D = \emptyset$; 
and we consider $C,D \neq \emptyset$.
Let ${CG}_1 = ((X_1,Y_1,E_1),\psi_1)$
and ${CG}_2 = ((X_2,Y_2,E_2),\psi_2)$.
Since $C = D$, by Lemma \ref{lem:lits}
$Lits(C) = Lits(D)$. From Defn. \ref{def:termset},
$(Terms(C) = Terms(D))$ iff $(Lits(C) = Lits(D))$.
From Defn. \ref{def:ClauseToGraph},
$(X_1 = X_2)$ iff $(Lits(C) = Lits(D))$ and
$(Y_1 = Y_2)$ iff $(Terms(C) = Terms(D))$.
If $(X_1 = X_2)$ and $(Y_1 = Y_2)$ then $(E_1 = E_2)$.
Since $h_x, h_y$ are bijections,
$\psi_1 = \psi_2$. That is,
${CG}_1 = {CG}_2$.
\end{myproof}

\begin{myproposition}[$ClauseToGraph$]
\label{prop:injection}
The function $ClauseToGraph: {\cal L}_{B,\Mu,d}$ $\rightarrow$ ${\cal {G}}$ (defined in  Defn.~\ref{def:ClauseToGraph})
is injective.
\end{myproposition}

\begin{myproof}
Let $C$ and $D$ in ${\cal L}_{B,\Mu,d}$, and
${CG}_1 = ClauseToGraph(C)$ and
${CG}_2 = ClauseToGraph(D)$. We need to show that
if ${CG}_1 = {CG}_2$ then $C = D$.

Let ${CG}_1$ = $(G_1,\psi_1)$ and
and ${CG}_2$ = $(G_2,\psi_2)$, where
    $G_1 = (X_1,Y_1,E_1)$ and $G_2 = (X_2,Y_2,E_2)$.
    Since ${CG}_1 = {CG}_2$,
    $(G_1,\psi_1) = (G_2,\psi_2)$. That is,
    $X_1 = X_2, Y_1 = Y_2$ and $\psi_1$ = $\psi_2$.
    Suppose
    $C \neq D$. Then, either there is some
    literal in $C$ that is not in $D$ or
    {\em vice versa\/}. Let $\lambda_i \in C$, and $\lambda_i \not \in D$. Let
    $\lambda_i$ be the corresponding
    literal in $\Lambda_C$, and
    $\lambda_i \not \in \Lambda_D$.
    Then since
    $C \in {\cal L}_{B,\Mu,d}$
    there must be
    at least one $\mu_i \in \Mu$ s.t.
    $(\lambda_i,\mu_i) \in Lits(C)$. Let $x = h_x((\lambda_i,\mu_i)) \in X_1$. 
    Since $h_x$ is a bijection, and $\lambda_i \not \in D$, there will be no other $\lambda$ and $\mu$ such that $h_x((\lambda,\mu))=x$. Hence $x \not \in X_2$.
    This is a contradiction, since $X_1 = X_2$.
    Similarly for $\lambda_i \in D$ and $\lambda_i \not \in C$.
\end{myproof}

\begin{myproposition}[Left-Inverse]
$ClauseToGraph(\cdot)$ has a left-inverse. 
\end{myproposition}

\begin{myproof}
We show that there is a
function $GraphtoClause: {\cal {G}} \rightarrow {\cal L}_{B,\Mu,d}$
s.t. for all $C \in \mathcal{L}_{B,\Mu,d}$, $GraphToClause(ClauseToGraph(C))=C$. 


\smallskip
\noindent
Let $CG = ClauseToGraph(C)$. 
So, $CG = (G,\psi)$, where
$G = (X,Y,E)$. For each $x_i \in X$:
\begin{enumerate}
    \item Let $L^+ = \{\lambda_i: x_i \in X, \psi(x_i) = (\lambda_i,\mu_i), \mu_i = modeh(\cdot)\}$;
    \item Let $L^- = \{\neg \lambda_i: x_i \in X, \psi(x_i) = (\lambda_i,\mu_i), \mu_i = modeb(\cdot)\}$
\end{enumerate}
Let $GraphToClause(CG) = C'$ where
$C' = L^+ ~\cup~ L^-$. We claim $C = C'$. 
Assume $C \neq C'$. Then there must be some literal
$l_i \in C$ s.t. $l_i \not \in C'$ (or {\em vice versa\/}). Let
the corresponding literal in $\Lambda_C$ be $\lambda_i$.
Since $C \in {\cal L}_{B,\Mu,d}$, there must be some
$\lambda\mu$-sequence (Defn. \ref{def:lmpair}) for $C$ s.t.
some $(\lambda_i,\mu_i) \in Lits(C)$ (Defn. \ref{def:litset}) and $h_x((\lambda_i,\mu_i)) \in X$ (Defn. \ref{def:ClauseToGraph}).
Then, by the construction above, $l_i \in C'$, which is a
contradiction. Suppose $l_i \in C'$ and $l_i \not \in C$. Then
there cannot be any $(\lambda_i,\mu_i)$ s.t.
$h_x((\lambda_i,\mu_i) \in X$. By construction, $l_i \not \in C'$,
which is a contradiction. Therefore there is no
$l_i \in C$ and $l_i \not \in C'$, or {\em vice versa\/}, and
$C = C'$.
\end{myproof}

\begin{myremark}
We note the following without formal proofs:
\begin{enumerate}[(i)]
    \item In Defn.~\ref{def:botgraph}, if there exists
    a unique $\bot_{B,\Mu,d}(e) \in {\cal L}_{B,\Mu,d}$
        then $BotGraph_{B,\Mu,d}(e)$ is unique.
        The proof follows from $BotGraph_{B,\Mu,d}(e)$ =
        $ClauseToGraph(\bot_{B,\Mu,d}(e))$.
    \item In Defn.~\ref{def:antgraph},
        $Antecededent:{\cal G} \rightarrow {\cal G}$ is well-defined. That is, if $Antecedent({CG}_1) \neq Antecedent({CG}_2)$ then
        ${CG}_1 \neq {CG}_2$. Again the proof follows from
        the contrapositive which is easily seen to hold. 
        Also, we note that $Antecedent$ is many-to-one,
        that is is possible that
        ${CG}_1 \neq {CG}_2$, and $Antecedent({CG}_1)$ =
        $Antecedent({CG}_2)$.
    \item In Defn.~\ref{def:ugraph}, $UGraph:{\cal G} \rightarrow {\cal G}$ is well-defined. That is,  if $UGraph({CG}_1) \neq UGraph({CG}_2)$     then ${CG}_1 \neq {CG}_2$. This follows from the
    contrapositive which is easily shown to
        hold (that is, 
        if ${CG}_1 = {CG}_2$ then $UGraph({CG}_1) = UGraph({CG}_2)$).
\end{enumerate}
\end{myremark}

\noindent
Finally, we relate the clausal
explanations found by some MDIE systems using the
ordering $\preceq_\theta$ 
defined over clauses in~\citep{plotkin1970note}.\footnote{
$C_1 \preceq_\theta C_2$ if there exists
some substitution $\theta$ s.t. $C_1 \theta \subseteq C_2$.
By convention $C_1$ is said to be more-general than
$C_2$, and $C_2$ is said to be more-specific than $C_1$.
It is known that if $C_1 \preceq_\theta C_2$, then $C_1\models C_2$.}
Given background knowledge $B$ and a clause $e$, we will
say a clause $C$ is a clausal explanation for $e$ if
$B \cup \{C\} \models e$.

\begin{myremark}[Relation to Clausal Explanations]
\label{rem:expl}

Let $\bot_{B,\Mu,d}(e)$ be
the ground most-specific definite-clause using MDIE.
Let $C$ be a clause  (not necessarily
ground). We show the following:
If $C\theta \subseteq \bot_{B,\Mu,d}(e)$ and
$C\theta \in {\cal L}_{B,\Mu,d}$, then
there exists a clause-graph ${CG}'$ s.t.
${CG}' \preceq_{cg} ClauseToGraph ($ $\bot_{B,\Mu,d}(e)$ and
${GraphToClause}({CG}') = C\theta$.

Denoting $\bot_{B,\Mu,d}(e)$ as $\bot(e)$ and
$ClauseToGraph(\bot_{B,\Mu,d})$ as ${CG}_{\bot(e)}$ the
relationships described in this remark is shown diagrammatically as:

\centerline{
    \includegraphics[width=0.4\textwidth]{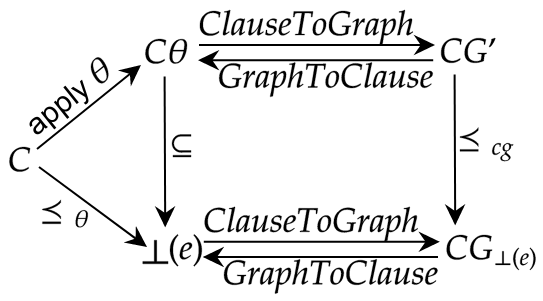}
}

Let $ClauseToGraph(\bot_{B,\Mu,d}(e)) = (G,\psi)$,
with $G = (X,Y,E)$. In the following, $Pos(l) = p$ if $l=\neg p$ is a negative literal, otherwise $Pos(l) = l$.
Consider the structure ${CG}' = (G',\psi')$, with $G' = (X',Y',E')$
obtained as follows.

\begin{enumerate}[(a)]
    \item $X' = \{x_i: x_i \in X,
    l_i \in C\theta$,
    $\lambda_i = Pos(l_i)$,
    $\psi(x_i) = (\lambda_i,\mu_i)\}$;
    \item $E' = \{(x_i,y_j): x_i \in X', (x_i,y_j) \in E\}$ $\cup$
            $\{(y_j,x_i): x_i \in X', (y_j,x_i) \in E\}$;
    \item $Y' = \{y_j: (x_i,y_j) \in E' ~\mathrm{or}~ (y_j,x_i) \in E'\}$;
    \item For $v \in X' \cup Y'$, $\psi'(v) = \psi(v)$
\end{enumerate}

It is evident that $G'$ is a directed bipartite graph, and
$\psi'$ is defined for every vertex in $G'$. So, ${CG}' \in {\cal G}$.
By construction, ${CG}'$ has the following properties:
(i) $X' \subseteq X$ and $Y' \subseteq Y$; 
(ii) $E' \subseteq E$; and
(iii) $\psi' \subseteq \psi$. 
Therefore ${CG}' \preceq_{cg} (G,\psi)$.
Since the vertices in $X'$ are obtained using only the literals in
$C\theta$, it follows that ${GraphToClause}(CG) = C\theta$.

Since $C\theta \subseteq \bot_{B,\Mu,d}(e)$, $C\theta \models \bot_{B,\Mu,d}(e)$. Further, since $C \preceq_\theta C\theta$,
$C \models C\theta$.
It follows that $B \cup C \models B \cup \bot_{B,\Mu,d}(e)$. Since
$B \cup \bot_{B,\Mu,d}(e) \models e$, then $B \cup C \models e$.
That is,
$C$ is a clausal explanation for $e$.

\end{myremark}

\section{Implementation Details on GNNs}
\label{app:gnnmaths}

We provide some basic mathematical details on how 
various graph convolutions and pooling operations are
implemented in the five different GNN variants considered
in our work. This section is meant for completeness only.
The reader should refer to the primary
sources of these implementations for a detailed
information on each of these approaches.

\subsection{Graph Convolution}
\label{app:graphconv}

For each graph convolution method, we only describe how the
$\mathsf{AGGREGATE}$-$\mathsf{COMBINE}$ procedure (as described
in Sec.~\ref{sec:gnn}) is implemented.

\subsubsection*{Variant 1: GCN}
\label{app:gnnvariant1}
Based on the spectral-based graph convolution as proposed by \cite{Kipf2017gcn}, this graph convolution uses a layer-wise (or iteration-wise) propagation rule for a graph with $N$ vertices as:
\begin{equation}
    \mathbf{H}^{(k)} = \sigma \left(\Tilde{D}^{-\frac{1}{2}}\Tilde{A}\Tilde{D}^{-\frac{1}{2}}\mathbf{H}^{(k-1)}\Theta^{(k-1)}\right)
    \label{eq:gcnconv}
\end{equation}
where, $H^{(k)} \in \mathbb{R}^{N \times D}$ denotes the matrix of vertex representations of length $D$, $\Tilde{A} = A + I$ is the adjacency matrix representing an undirected graph $G$ with added self-connections, $A \in \mathbb{R}^{N \times N}$ is the graph adjacency matrix, $I_N$ is the identity matrix, $\Tilde{D}_{ii} = \sum_{j}\Tilde{A}_{ij}$, and $\Theta^{(k-1)}$ is the iteration-specific trainable parameter matrix, $\sigma(\cdot)$ denotes the activation function e.g. $\mathrm{ReLU}(\cdot) = \max(0,\cdot)$, $\mathbf{H}^{(0)} = \mathbf{X}$, $\mathbf{X}$ is the matrix of feature-vectors of the vertices, where each vertex $i$
is associated with a feature-vector $X_i$.

\subsubsection*{Variant 2: $k$-GNN}
\label{app:gnnvariant2}
This graph convolution passes
messages (vertex feature-vectors) directly between subgraph
structures inside a graph~\citep{morris2019weisfeiler}. 
At iteration $k$, the feature representation of a vertex is computed by using
\begin{equation}
    h^{(k)}_{u} = \sigma \left( h^{(k-1)}_{u}\cdot \Theta^{(k)}_1 + \sum_{v \in \mathcal{N}(u)}{h^{(k-1)}_{v}\cdot \Theta^{(k)}_2} \right)
    \label{eq:graphconv}
\end{equation}
where, $h^{k}_{u}$ denotes the vertex-representation of a vertex $u$ at iteration $k$, $\mathcal{N}$ denotes the 
neighborhood function,
$\sigma$ is a non-linear transfer function applied component wise to the function argument, $\Theta$s are the layer-specific
learnable parameters of the network.

\subsubsection*{Variant 3: GAT}
\label{app:gnnvariant3}
This variant is based on aggregating information from
neighbours with attention. This approach is 
popularly known as Graph Attention Network (GAT:~\citep{velickovic2018graph}). This network assumes that the contributions of neighboring vertices to the central vertex are not pre-determined which is the case in the Graph Convolutional Network~\citep{Kipf2017gcn}. This adopts attention mechanisms to learn the relative weights between two connected vertices. The graph convolutional operation at iteration $k$ is thereby defined as:
\begin{equation}
    h^{(k)}_{u} = \sigma\left(\sum_{v \in \mathcal{N}(u) \cup u}{ \alpha_{uv}^{(k)}\Theta^{(k)}h^{(k-1)}_{u}} \right)
\end{equation}
where, $h^{k}_{u}$ denotes the vertex-representation of a vertex $u$ at iteration $k$;
$h^{(0)}_u = X_u$ (the initial feature-vector
associated with a vertex $u$). The connective strength between the vertex $u$ and its neighbor vertex $v$ is called attention weight, which is defined as
\begin{equation}
    \alpha^{(k)}_{uv} = \mathrm{softmax}\left(\mathrm{LeakyReLU}\left( a^\mathsf{T}\left[\Theta^{(k)}h^{(k-1)}_u \mathbin{\|} \Theta^{(k)}h^{(k-1)}_v \right]\right)\right)
\end{equation}
where, $a$ is the set of learnable parameters of a single layer feed-forward neural network,
$||$ denotes the concatenation operation.

\subsubsection*{Variant 4: GraphSAGE}
\label{app:gnnvariant4}
This graph convolution is based on inductive representation learning on large graphs~\citep{hamilton2017inductive},
which is primarily used to generate low-dimensional vector representations for vertices. 
It adopts two steps: First, it samples a neighbourhood
vertices of a vertex; Second, aggregate the feature-information from these sampled vertices. 
GraphSAGE is used to found to be very useful for 
graphs with vertices associated with rich feature-vectors.
The following is an iterative update of the vertex
representations in a graph:
\begin{equation}
    h^{(k)}_{u} = \sigma \left( h^{(k-1)}_{u}\cdot \Theta^{(k)}_1 + \frac{1}{|\mathcal{N}(u)|} \sum_{v \in \mathcal{N}(u)}{h^{(k-1)}_{v}\cdot \Theta^{(k)}_2} \right)
    \label{eq:sageconv}
\end{equation}
where, $h^{k}_{u}$ denotes the vertex-representation of a vertex $u$ at iteration $k$,
$\sigma$ is a non-linear transfer function applied component wise to the function argument, 
$\mathcal{N}$ denotes the 
neighborhood function,
$\Theta$s are the layer-specific
learnable parameters of the network.

\subsubsection*{Variant 5: ARMA}
\label{app:gnnvariant5}
This graph convolution is inspired by the auto-regressive moving average (ARMA) filters that are considered to be more robust than polynomial
filters~\citep{bianchi2021graph}. The ARMA graph convolutional operation is defined as:
\begin{equation}
    \mathbf{H}^{(k)} = \frac{1}{M} \sum_{m=1}^{M}{\mathbf{H}^{(K)}_m}
\end{equation}
where, $\mathbf{H}^{k}$ denotes the vertex-representation
matrix at iteration $k$,
$M$ is the number of parallel stacks, 
$K$ is the number of layers; and $\mathbf{H}^{(K)}_m$ is recursively defined as
\begin{equation}
    \mathbf{H}^{(k+1)}_{m} = \sigma\left(\hat{L}\mathbf{H}^{(k)}_{m}\Theta^{(k)}_2 + \mathbf{H}^{(0)}\Theta^{(k)}_2 \right)
\end{equation}
where, $\sigma$ is a non-linear transfer function,
$\hat{L} = I - L$ is the modified Laplacian. The $\Theta$ parameters are learnable parameters. 

\subsection{Graph Pooling}
\label{app:graphpool}

Graph pooling methods apply the idea of downsampling mechanisms to graphs.
This operation allows us to obtain refined graph representations at each layer. The primary aim of 
including a graph pooling operation after each
graph convolution is that this operation can 
reduce the graph representation while ideally preserving important structural information. 
In this work, we use a recently proposed graph pooling method
based on self-attention~\citep{lee2019self}.
This method uses the graph convolution defined in Eq.~\eqref{eq:gcnconv} to obtain a self-attention score as given in Eq.~\ref{eq:sagpool} with the trainable parameter replaced by $\Theta_{att} \in \mathbb{R}^{N\times 1}$, which is a set of trainable parameters in the pooling layer.
\begin{equation}
    Z = \sigma \left(\Tilde{D}^{-\frac{1}{2}}\Tilde{A}\Tilde{D}^{-\frac{1}{2}}\mathbf{X}\Theta_{att}\right)
    \label{eq:sagpool}
\end{equation}
Here, $\sigma(\cdot)$ is the activation function e.g. $\tanh$.

\subsection{Hierarchical Graph Pooling}

The graph-convolution and graph-pooling operations
described in the preceding two subsections allows a 
GNN are concerned with construction of vertex-representations. To deal with the problem of
graph classification (as is the case in this work),
we need to represent an input graph as a ``flattened''
fixed-length feature-vector that can then be used with a standard
fully-connected multilayer neural network (e.g.
Multilayer Perceptron) to produce a class-label.
To construct this graph-representation (mostly, 
a dense real-valued feature-vector, also called a \textit{graph-embedding}), we use hierarchical graph-pooling method proposed by \cite{cangea2018towards}.
This method is implemented with two operations: 
(a) global average pooling, that
averages all the learnt vertex representations
in the final (readout) layer; 
(b) augmenting the representation obtained in (a)
with the representation obtained using global max pooling,
that seek to obtained the most relevant information
and could strengthen the graph-representation.
The term ``hierarchical'' refers to the fact that
the above two operations (a) and (b) are carried
out after each conv-pool block in the GNN (refer
Fig.~\ref{fig:botgnn_arch}). The final 
graph representation is an aggregate of all the
layer-wise representations by taking their sum. 

The output graph after each conv-pool block can be
represented by a concatenation of the global
average pool representation and the global max pool
representation as:
\begin{equation}
    H_G^{(k)} = \mathrm{avg}(\mathbf{H}^{(k)})~||\max(\mathbf{H}^{(k)})
    \label{eq:h_g}
\end{equation}
where, $H_G^{k}$ denotes the graph-representation
at iteration $k$; $\mathbf{H}^{(k)}$ denotes the matrix
of vertex-representations after conv-pool operations at
iteration $k$ as mathematically described in Sec.~\ref{app:graphconv}
and Sec.~\ref{app:graphpool};
$\mathrm{avg}$ and $\max$ denote the average and 
max operations, which are computed as follows:
\begin{equation}
    \mathrm{avg}(\mathbf{H}^{(k)}) = \frac{1}{N}\sum_{i=1}^{N}{\mathbf{H}^{(k)}_i}
\end{equation}
\begin{equation}
    \max(\mathbf{H}^{(k)}) = \max_{i=1}^{N}{\mathbf{H}^{(k)}_i}
\end{equation}
Here, $\mathbf{H}^{k}_i$ denotes the representation for the $i$th vertex of the graph; 
$N$ is the number of nodes in the graph.

The final fixed-length representation after iteration $K$ for the whole
input graph is then computed by the element-wise sum, denoted as $\oplus$, 
of these intermediate graph-representations in Eq.~\ref{eq:h_g}:
\begin{equation}
    H_G^{(K)} = \oplus_{k=1}^{K} H^{(k)}_G
\end{equation}
In our present work, $K=3$ 
(since, we use 3 conv-pool blocks in our GNN). 
The graph-representation $H^{(K)}_G$
is then input to a multilayer perceptron as 
described in Sec.~\ref{sec:meth}.

\section{Application Details}
\label{app:appdetails}

\subsection{Mode-Declarations}

We use the ILP engine, Aleph~\citep{srinivasan2001aleph} to
construct the most-specific clause for a relational
data instance given background-knowledge, mode
specifications and a depth.
The mode-language used for our main experiments
in the paper is given below:
\begin{verbatim}
    :- modeb(*,bond(+mol,-atomid,-atomid,#atomtype,#atomtype,#bondtype)).
    :- modeb(*,has_struc(+mol,-atomids,-length,#structype)).
    :- modeb(*,connected(+mol,+atomids,+atomids)).
    :- modeb(*,fused(+mol,+atomids,+atomids)).
\end{verbatim}
The `\#'-ed arguments in the mode declaration 
refers to type, that is, $\mathtt{\#atomtype}$ refers to the type of atom, $\mathtt{\#bondtype}$ refers to the type of bond, and $\mathtt{\#structype}$ refers to the 
type of the structure (functional group or ring) associated with the molecule. 

\subsection{Logical Representation of the Data and Background Knowledge}

Data in the primary set of experiments are molecules.
At the lowest level, the atoms and bonds in
each molecule are represented as a set of
ground definitions of the $\mathtt{bond}/6$ predicate.
Thus {\tt bond(m1,27,24,o2,car,1)}
denotes that in instance {\tt m1}
there is an oxygen atom (id 27), and a carbon atom (id 24) connected
by a single bond ({\tt car} denotes a carbon atom in an aromatic ring).
Definitions of functional-groups and ring-structures are written as
clausal definitions that use the definitions of this low-level
$\mathtt{bond}/6$ predicate. There are about 100 such higher-level definitions. The result of inference about the presence of functional
groups and rings is pre-compiled for efficiency. For example: 
\begin{verbatim}
    functional_group(m1,[27],1,oxide).
    ring(m1,[25,28,30,29,26,23],6,benzene_ring).
\end{verbatim}

\noindent
Here, the 1 and 6 denote number of atoms involved in the group
or ring.
\noindent
Access to groups and rings in a molecule uses the
$\mathtt{has\_struc}/4$ predicate:

\begin{verbatim}
    has_struc(Mol,Atoms,Length,Type):-
        ring(Mol,Atoms,Length,Type).
    has_struc(Mol,Atoms,Length,Type):-
        functional_group(Mol,Atoms,Length,Type).
    ...
\end{verbatim}

\noindent
In addition to functional groups and rings, there are predicates
for computing if structures are connected or fused.
An example of the most-specific clause for a data
instance is:
\begin{verbatim}
    class(m411,pos):-
        bond(m411,18,13,c3,o2,1),
        bond(m411,13,18,o2,c3,1),
        bond(m411,12,16,nar,nar,ar),...,
        has_struc(m411,[5,7,10,6,4],5,pyrrole_ring),
        has_struc(m411,[7,11,16,12,8,5],6,pyridazine_ring),...,
        connected(m411,[15],[9,14,13]),
        connected(m411,[15],[7,11,16,12,8,5]),...,
        fused(m411,[7,11,16,12,8,5],[5,7,10,6,4]),...,
        lteq(1,1),lteq(3,3),...,
        gteq(1,1),gteq(3,3),....
\end{verbatim}

\subsection{Propositionalisation Experiments}

In a propositionalisation approach~\citep{lavravc1991learning}, 
each data instance is represented using a Boolean vector
of 0's and 1's, depending on the value of propositions
(constructed manually or automatically) for the data instance
(the value of the $i^{\mathrm{th}}$ dimension is 0 if
the $i^{\mathrm{th}}$ proposition is false for the data-instance
and 1 otherwise). The resulting dataset is then 
used to construct an MLP model. The following details are relevant:
\begin{itemize}
    \item The MLP is implemented using Tensorflow-Keras~\citep{chollet2015keras}.
    \item The number of layers in MLP is tuned using
    a validation-based approach. The parameter grid for
    number of hidden layers is: $\{1,2,3,4\}$.
    \item Each layer has fixed number of neurons: 10.
    \item The dropout rate is 0.5. We apply dropout~\citep{srivastava2014dropout}
    after every layer in the network except the
    output layer. 
    \item The activation function used in each hidden layer
    is $\mathtt{relu}$.
    \item The training is carried out using the Adam optimiser~\citep{kingma2014adam}
    with learning rate 0.001.
    \item Additionally, we use early-stopping~\citep{prechelt1998early} to 
    control over-fitting during training.
\end{itemize}

\noindent
For the DRM, propositions simply denote whether any specific
relation in the background knowledge is true or false for 
the data instance. BCP~\citep{francca2014fast} constructs propositions using the
most-specific clauses returned by the ILP system Aleph
given the background-knowledge $B$, modes $\Mu$ and
depth-limit $d$. For the construction of
Boolean features using BCP, we use the
code available at \citep{vakker2020cilp}.

\subsection{Experiments with ILP Benchmarks}

The seven datasets are taken from~\citep{srinivasan2003empirical}. These datasets
are some of the most popular benchmark datasets to evaluate various techniques within ILP studies.
For the construction of
BotGNNs the following details are relevant:
\begin{itemize}
    \item There is a BK for each dataset. 
    \item There are 10 splits for each dataset. Therefore,
    for each test-split we construct BotGNNs (all 5 variants),
    using 8 of rest splits as training set and the remaining
    1 split as a validation set.
    \item Since these datasets are small (few hundreds of data instances), we could manage
    to perform some hyperparameter tuning for construction of our BotGNNs. The parameter grids for this are: $m: \{8, 16, 32, 64, 128\}$; batch-size: $\{16, 32\}$; learning rate: $\{0.0001, 0.0005, 0.001\}$.
    \item Other details are same as described in the 
    main BotGNN experiments.
    \item In Section \ref{sec:additional} we
        report the test accuracy from the best performing
        BotGNN variant. 
\end{itemize}

\end{document}